\documentclass[twoside]{article}
\usepackage{natbib}
\usepackage{graphicx}
\usepackage{booktabs}
\usepackage{multirow}
\usepackage{amsmath, amsthm, amssymb, amsfonts}
\usepackage{hyperref}
\usepackage{subcaption}
\usepackage[capitalize,noabbrev]{cleveref}
\usepackage[preprint]{aistats2027}

\newcommand{\R}{\mathbb{R}}
\newcommand{\bA}{\mathbf{A}}
\newcommand{\bC}{\mathbf{C}}
\newcommand{\bR}{\mathbf{R}}
\newcommand{\bS}{\mathbf{S}}
\newcommand{\bX}{\mathbf{X}}
\newcommand{\bx}{\mathbf{x}}
\newcommand{\bY}{\mathbf{Y}}
\newcommand{\bZ}{\mathbf{Z}}
\newcommand{\bz}{\mathbf{z}}
\newcommand{\bs}{\mathbf{s}}
\newcommand{\bc}{\mathbf{c}}
\newcommand{\bh}{\mathbf{h}}
\newcommand{\br}{\mathbf{r}}
\newcommand{\one}{\mathbf{1}}

\newtheorem{lemma}{Lemma}
\newtheorem{theorem}{Theorem}

\DeclareMathOperator*{\argmin}{arg\,min}
\DeclareMathOperator{\softmax}{softmax}
\DeclareMathOperator{\sparsemax}{sparsemax}

\begin{document}

%

%
\runningauthor{Wedenborg, Nørskov, Dorszewski, Wickstrøm, Mørup}

\twocolumn[

\aistatstitle{The Polytopal Neural Network}

\aistatsauthor{
    \begin{tabular}{c}
        A. Emilie J. Wedenborg \\
        \normalfont Technical University of Denmark
    \end{tabular}
    \And
    \begin{tabular}{c}
        Anders V. Nørskov \\
        \normalfont Technical University of Denmark
    \end{tabular}
    \And
    \begin{tabular}{c}
        Teresa Dorszewski \\
        \normalfont Technical University of Denmark
    \end{tabular}
    \AND
    \begin{tabular}{c}
        Kristoffer Wickstrøm \\
        \normalfont UiT The Arctic University of Norway
    \end{tabular}
    \And
    \begin{tabular}{c}
        Morten Mørup \\
        \normalfont Technical University of Denmark
    \end{tabular}
}

\aistatsaddress{}
]

\author{
  \authorname{A. Emilie J. Wedenborg}$^{1}$ \quad
  \authorname{Anders V. Nørskov}$^{1}$ \quad
  \authorname{Teresa Dorszewski}$^{1}$ \\
  \authorname{Kristoffer Wickstrøm}$^{2}$ \quad
  \authorname{Morten Mørup}$^{1}$ \\
  $^{1}$ Technical University of Denmark \quad
  $^{2}$ The Arctic University of Norway \\
  \texttt{\{aejew,tksc,aveno,mmor\}@dtu.dk} \quad
  \texttt{kristoffer.k.wickstrom@uit.no}
}

\begin{abstract}
  Understanding how deep neural networks process information remains a central challenge. Existing interpretability methods often compromise structural fidelity, rely on prespecified corpora, or explain models post-hoc. We propose Polytopal Neural Networks (PNNs), a framework that extracts distinct layer-wise aspects by enforcing a polytope-based structure that is used directly in subsequent information processing. We scale our approach using learned corpus representations and an amortized simplex inference procedure and highlight how the framework also gives a direct route to vector quantized (VQ) training. In PNNs, observations are explicitly described by their alignment with layer-specific aspects. Empirical results show that imposing polytopal constraints on neural network representations preserves meaningful structures in the latent space with minimal degradation in performance, favorable compressed representations when compared to VQ representations in unsupervised learning, while also providing a performant new approach to VQ deep learning training. Our findings suggest that deep networks can enforce interpretable polytope-based representations, offering a principled path toward more transparent AI systems with minimal performance compromise.
\end{abstract}

\begin{figure}[!ht]
    \centering
    \includegraphics[width=\linewidth]{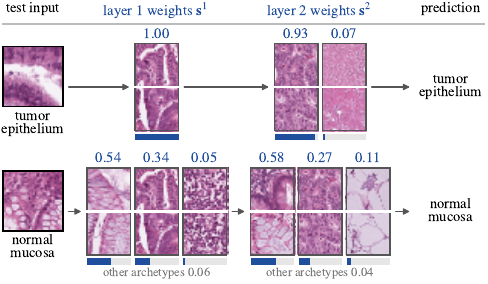}
    \caption{Two PathMNIST test images traced through two PNN layers ($K=10$). Each layer passes a convex combination of archetypes to the next layer. Numbers denote mixture weights; each archetype is illustrated by its two nearest training images.
    }
    \label{fig:PNN_example}
\end{figure}
\section{Introduction}

The increasing deployment of deep learning systems in scientific, industrial, and societal contexts has intensified the demand for methods that make their internal computations transparent and trustworthy \citep{LONGO2024102301}. Yet, many existing approaches fail to explain how neural networks structure and transform information across layers. A substantial body of explainable AI (XAI) research has focused on post-hoc analyses of input-output behavior, including saliency-based attribution methods and activation visualizations \citep{bach2015pixel, selvaraju2017grad,uendes2025start}. Although such techniques can identify influential input regions, they offer limited insight into the internal mechanisms directing the representation learning within the network. This limitation is especially noticeable given consistent empirical evidence that deep networks develop structured, hierarchical representations, where early layers encode low-level attributes such as color and texture, while deeper layers assemble these primitives into higher-order concepts and object-level abstractions \citep{bau2017network, dorszewski2025colors, vielhaben2024beyond}. Despite their central role in learning and inference, these latent representations remain largely unconstrained and difficult to interpret, restricting their usefulness for mechanistic understanding, model validation, and scientific discovery.
 
We introduce Polytopal Neural Networks (PNNs), a class of architectures in which latent representations are explicitly constrained to lie on a polytope by use of simplex constraints (i.e., imposing non-negativity and sum-to-one constraints). This induces a structured, interpretable latent geometry in which each representation is expressed as a convex combination of learned archetypes defining the corners of the polytope (Figure~\ref{fig:PNN_example}). Crucially, this constraint is enforced during training and shapes the network’s internal computations, rather than serving as a post-hoc explanation mechanism \citep{AA_NLDL2026}. PNNs are related to corpus-based explanation methods that express predictions as mixtures of exemplars \citep{Crabbe2021ExplainingExamples}. However, these approaches rely on a fixed, user-defined corpus and aim to explain individual predictions after training. In contrast, PNNs learn the archetypal corpus end-to-end and enforce simplex constraints directly on the latent space, ensuring that predictions are intrinsically consistent with the interpretable polytope representations imposed at every layer.

We use interpretability to mean the extent to which a model’s internal representations and computations can be understood in terms of identifiable components and their contributions. An explanation is intrinsic when those components are part of the computation producing the output, rather than estimated afterward. It is local when it describes an individual input and global when it describes structure shared across inputs. PNNs provide intrinsic local explanations through each input’s mixture weights over archetypes, and global structure through the shared archetypes that define each layer’s polytope. This guarantees a faithful description of the constrained representations.

PNN’s use of archetypes as reference profiles is motivated by geometric accounts of cognition based on similarity and betweenness \citep{1642090.1642143}, and by evidence that contrastive categorization can favor extreme, idealized representatives along distinguishing dimensions \citep{Tyler2010}. This framing connects the PNN to classical Archetypal Analysis (AA) \citep{Cutler1994ArchetypalAnalysis, Mrup2012ArchetypalMining}, which models data as mixtures of extremal points lying on the convex hull. AA based models are well known for their interpretability, as both the archetypes and the mixture coefficients allow intuitive explanations. Integrating AA constraints into deep architectures, PNNs bridge the gap between interpretable, prototype-based modeling and high-capacity representation learning, ensuring the results remain strictly grounded in the original data.

The proposed framework also relates to vector quantization (VQ) methods, most notably vector quantized variational autoencoders (VQ-VAEs) \citep{OordVQVAE}, which replaces continuous latent manifolds with finite representations to promote compression, disentanglement, and efficient generative modeling. Both VQ-VAE and PNNs reduce latent complexity by constraining representations to a limited set of learned components. However, their mechanisms differ: VQ-VAE relies on hard nearest-neighbor assignments, exponential moving average (EMA) updates, and auxiliary optimization heuristics such as the straight-through estimator and commitment losses \citep{OordVQVAE}, whereas PNNs employ continuous convex combinations of archetypes, avoiding discrete assignment operations while preserving a compact and interpretable latent structure.

In summary, PNNs combine the strengths of Archetypal Analysis, Vector Quantization, and deep representation learning. 
Each layer decomposes activations into convex mixtures of learned archetypes, forming a sequence of sequential polytopes that jointly explain and reconstruct the model’s internal representations. 
This polytopal structure enforces geometric consistency and provides an intrinsic explanation of each prediction through archetypal coefficients. 

\begin{figure*}[!ht]
    \centering
    \includegraphics[width=\linewidth]{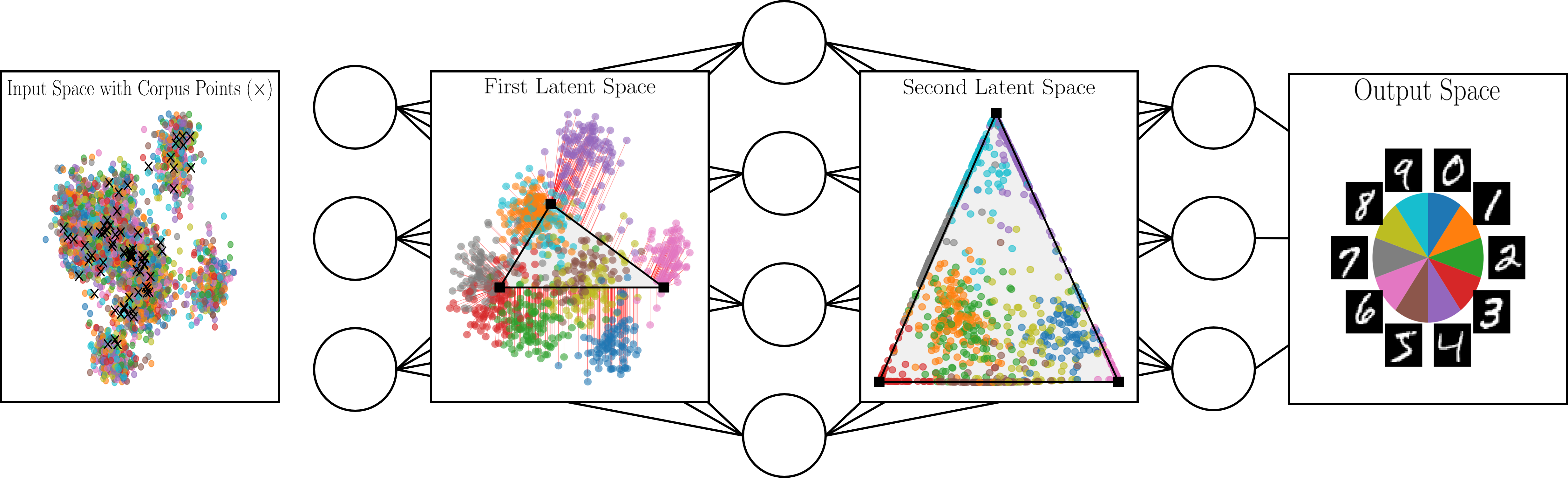}
    \caption{\textbf{The Polytopal Neural Network: }Points in the input space (left) are used to learn data driven corpus points ($\times$)  used to span the polytopes in the layers of the PNN. Each layer maps the observations into a low-dimensional simplex-structured latent space (polytope), where convex combinations of the encoded corpus points define the layer specific archetypal vertices. A subsequent latent transformation refines the geometric organization forming a refined polytope. Finally, points are decoded into the output space (right), where the resulting representation exhibits improved separation and structure. Colored points indicate different data groups, and black vertices denote learned archetypes.}
    \label{fig:fig1}
\end{figure*}

Specifically, our main contributions are: \\

\textbf{(i) Polytopal representation learning.} A layer-wise framework that constrains latent features to lie within learned polytopes, with AA constrained representations enforced exactly at every constrained layer.

\textbf{(ii) Integrated interpretability.} The coordinates used to explain a representation are also the coordinates that the next layer receives.

\textbf{(iii) Scalable inference.} A batched learning framework with (a) a compact corpus learned jointly with the network and updated from mini-batches and (b) amortized simplex inference.

\textbf{(iv) A direct optimization route for VQ.} Replacing the simplex with one-hot coordinates reduces the PNN to a VQ model whose codebook is grounded in the corpus and is trained without EMA updates, a commitment loss, or a straight-through estimator.

\section{Related Work}
Archetypal Analysis (AA), introduced by \cite{Cutler1994ArchetypalAnalysis}, has been incorporated into deep learning frameworks, including sparse autoencoder–based models \citep{pmlr-v267-fel25a} and other deep architectures \citep{DeepArchetypalAnalysis, DeepAAExtremal, milite2025midaa, wieser2025revisiting}. Most existing non-linear AA approaches place archetypes only at a single latent bottleneck of an autoencoder, and several enforce the archetypal structure only through a penalty \citep{DeepArchetypalAnalysis, DeepAAExtremal, wieser2025revisiting}, so the archetypes act as a geometric prior rather than a strict constraint, limiting interpretability and consistency across layers.


Early work such as AAnet proposed learning nonlinear archetypal spaces using deep autoencoders that map data into a latent simplex, demonstrating improved recovery of extremal structure in nonlinear domains \citep{vanDijk2019FindingAA}. 

Recent research highlights AA as a tool for explainable representation learning, from stabilizing sparse autoencoders \citep{pmlr-v267-fel25a} to identifying task-relevant concepts in vision transformers \citep{fel2025rabbithulltaskrelevantconcepts}. Post hoc probing confirms that stable archetypes exist within pretrained networks \citep{AA_NLDL2026} but leaves the network unchanged. Deep AA architectures, in turn, place archetypes at a single bottleneck. There remains a need for architectures that enforce simplex-constrained representations throughout the network depth with explanations fully consistent with the networks' information processing.

An alternative paradigm learns prototypical summaries jointly with the model. Prototype- and clustering-based approaches encourage representation space to organize around learned centroids or parts, facilitating concept-level explanations \citep{chen2024neural, liang2023clusterfomer,wang2023visual, zhou2024prototype}. ProtoPNet \citep{chen2019looks} assigns prototypes to classes, while PIP-Net \citep{Nauta_2023_CVPR} learns prototype–class associations through supervised classification after self-supervised pretraining. The PNN instead constrains intermediate representations to convex combinations of $K$ archetypes without class assignments, using only the task loss. This constraint applies at multiple classifier depths, in unlabeled autoencoders, and to tokens from pretrained backbones. We therefore compare against the same unconstrained architecture and other representation bottlenecks: projection layers, vector quantization, and the Dirichlet VAE. Centroid-oriented representations emphasize typical instances and are effective for class- or part-centered interpretations. Notably, centroid-based representations are also explored in VQ-VAEs \citep{OordVQVAE}. By contrast, AA targets extremal data geometry, decomposing space into vertices of a convex polytope and expressing each point as a convex mixture of these vertices \citep{Cutler1994ArchetypalAnalysis, hastie2017elements, Mrup2012ArchetypalMining}. This shift from central tendency (centroids) to extremal representation (archetypes) yields complementary interpretive advantages where archetypes capture boundary phenomena and diverse modes of variation that centroids can obscure.

PNNs more closely resemble the Dirichlet VAE \citep{joo2020dirichlet} in which the latent representation is projected onto the standard simplex with an associated Dirichlet prior imposed to characterize the continuum within the standard simplex. The PNN notably differs here by imposing a learned polytope as opposed to a fixed projection onto the standard simplex. The use of subspace projections of latent representations has also been widely adopted in deep learning using linear projection layers, see also \citep{hawkins2018projection} and the recent generalization to arbitrary higher-order feature spaces \citep{morimoto2025tensorprojection}. Whereas these procedures can produce lower-dimensional subspaces they do not promote explainable polytopal latent representations.

\section{Methods}\label{sec:method}
\paragraph{Notation.} Data and representations are observations as rows throughout. $\bX \in \R^{N\times M}$ holds $N$ observations. At layer $\ell$, $\bZ^\ell \in \R^{N \times d_\ell}$ holds the representations before projection, $\bC^\ell \in \R^{K\times N}$ is row-stochastic, and $\bS^\ell \in \R^{N\times K}$ is row-stochastic. A row-stochastic matrix has non-negative entries and rows that sum to one. The archetypes are $\bA^\ell = \bC^\ell \bZ^\ell \in \R^{K\times d_\ell}$ and the projected representations are $\bR^\ell = \bS^\ell \bA^\ell$.

\subsection{Archetypal Analysis and the Polytopal Neural Network}

AA approximates the convex hull of the data by a polytope with $K$ vertices, the archetypes, which are convex combinations of the data \citep{Mrup2012ArchetypalMining}. It solves
\begin{equation}\label{eq:AAClassic}
\begin{aligned}
&\min_{\bC,\bS} \|\bX-\bS\bC\bX\|_F^2 \\
&\text{s.t. } \bc_k \ge 0,\ \bc_k\one = 1\ \forall k,\ \bs_n \ge 0,\ \bs_n\one = 1\ \forall n,
\end{aligned}
\end{equation}
where $\bc_k$ and $\bs_n$ are the $k$th row of $\bC$ and the $n$th row of $\bS$. The rows of $\bC\bX$ are the archetypes and $\bS$ projects each observation onto the polytope they span. Archetypes are therefore idealized, data-grounded extremes, and each observation has a transparent decomposition in terms of these extremes.

A deep network applies $L$ layers $f^\ell_{\theta^\ell}$ and is trained by minimizing $\mathcal{L}(\bY, f_\theta(\bX))$ with $f_\theta = f^L_{\theta^L}\circ\cdots\circ f^1_{\theta^1}$. Supervised learning uses labels as $\bY$, and an autoencoder uses the inputs themselves. The PNN inserts a projection operator $\Delta^\ell_{\bC^\ell}$ after the chosen layers,
\begin{equation}\label{eq:main_obj_ext}
\begin{aligned}
f_\theta(\bX) = f^{L}_{\theta^{L}}&\circ \big[\Delta^{L-1}_{\bC^{L-1}} \circ f^{L-1}_{\theta^{L-1}}\big] \\
&\circ \cdots \circ \big[\Delta^{1}_{\bC^{1}} \circ f^{1}_{\theta^{1}}\big](\bX),
\end{aligned}
\end{equation}
and learns $\{\theta^\ell, \bC^\ell\}$ subject to $\bC^\ell$ being row-stochastic. With $\bZ^\ell$ the output of layer $\ell$ before projection, $\Delta^\ell_{\bC^\ell}(\bZ^\ell) = \bS^\ell\bC^\ell\bZ^\ell$, where
\begin{equation}\label{eq:PolytopalProjection}
\bS^\ell = \argmin_{\bS} \|\bZ^\ell-\bS\bC^\ell\bZ^\ell\|_F^2 \ \text{ s.t. } \ \bs_n \ge 0,\ \bs_n\one = 1 \ \forall n.
\end{equation}
This is a convex quadratic program (QP) per sample. We parameterize $\bC^\ell = \softmax(\tilde\bC^\ell)$ row-wise with unconstrained logits $\tilde\bC^\ell$ and train all parameters by back-propagation, treating $\bS^\ell$ as fixed when differentiating with respect to $\bC^\ell$ (Appendix~\ref{app:smo}). An exact solve by an active-set method costs up to $\mathcal{O}(K^3)$ per sample. Sequential minimal optimization (SMO) \citep{AABinary} reduces a full pass to $\mathcal{O}(K^2)$ by updating two coordinates at a time in closed form. Storing $\bX$, every $\bZ^\ell$, and $\bC^\ell$ still costs $\mathcal{O}(NM + N\sum_l d_\ell+NLK)$ in memory.

\subsection{Scaling the PNN}
\paragraph{Scaling Memory Usage by Batched Streaming Updates of Data-Driven Corpus:}\label{subsection:corpus_updates}

We decouple the archetypes from the full dataset by learning a corpus of $N^{c} \ll N$ points, $\bX^{c} = \bC^{c}\bX$ with row-stochastic $\bC^{c}\in\R^{N^c\times N}$. The corpus is encoded by the same layers as the data, $\bZ^{c,\ell}$, and defines the archetypes $\bA^\ell = \bC^\ell\bZ^{c,\ell}$ with $\bC^\ell\in\R^{K\times N^c}$. Let $\tilde c_{ki}$ be the logits of corpus row $k$. For numerical stability we subtract the row maximum of all $N$ observations, $w_{ki} = \exp(\tilde c_{ki} - \max_j \tilde c_{kj})$, which leaves the softmax unchanged. For a mini-batch $B$ and its complement $\neg B$, corpus row $k$ is
\begin{equation}\label{eq:streaming_XC}
\bx^{c}_k = \frac{\sum_{i\in B} w_{ki}\bx_i + \sum_{i\in \neg B} w_{ki}\bx_i}{\sum_{i\in B} w_{ki} + \sum_{i\in \neg B} w_{ki}}.
\end{equation}
The batch sums carry gradients. The non-batch sums are computed from the current logits without gradients, in chunks. A row outside the batch therefore receives no gradient on that step, but any update it received while in an earlier batch is reflected the next time the statistics are recomputed. Eq.~\eqref{eq:streaming_XC} is the exact weighted average at every step, and gradient memory scales with $|B|$ and $N^c$ rather than with $N$ resulting in a memory usage of $\mathcal{O}(N^c|B|+N^cM+|B|M + N^c\sum_l d_\ell+N^cLK)$.


\paragraph{Amortized simplex inference.}\label{sec:amortized}
Although exact inference solves Eq.~\eqref{eq:PolytopalProjection} for every sample in every forward pass at cost of $\mathcal{O}(K^3)$ reduced by SMO to $\mathcal{O}(K^2)$ this can still be prohibitive. We therefore instead learn a map $g_\phi^\ell$ on the archetype correlations $\bh_n = \bz_n (\bA^\ell)^\top \in \R^K$,
\begin{equation}\label{eq:amortizer}
\hat\bs_n = \softmax\big(g_\phi^\ell(\bh_n)\big), \qquad \hat\br_n = \hat\bs_n\bA^\ell.
\end{equation}
The input of $g^\ell_\phi$ is a layer normalization of $\bh_n$ together with $\log \mathrm{std}(\bh_n)$, because the projection depends on the scale of $\bh_n$, which layer normalization removes. The task loss is computed through $\hat\br_n$, but $g^\ell_\phi$ is never trained on it. It is trained only on the projection residual $\|\bz_n - \hat\bs_n\bA^\ell\|^2$ with $\bz_n$ and $\bA^\ell$ detached, so it cannot deform the encoder, the head, the corpus, or the archetypes. It is fitted before the archetypes are trained and refit every ten epochs as the polytope moves. At test time the amortized coordinates can be refined by $K$ rounds of batched SMO warm-started at $\hat\bs_n$ (Appendix~\ref{app:smo}).

\subsection{Theoretical Properties}\label{sec:theory}

For a fixed polytope, let $A\in\R^{K\times d}$ hold the archetypes as columns, $G=A A^\top$, $h=A z$, and $\Delta^{K-1}=\{s\in\R^K\mid s\ge0,\ \one^\top s=1\}$. Expanding $\tfrac12\|z-A^{\top}s\|^2$ and dropping the term independent of $s$ gives
\begin{equation}\label{eq:qp}
\Phi(h)=\argmin_{s\in\Delta^{K-1}}
\tfrac12 s^\top Gs-h^\top s.
\end{equation}

\begin{lemma}\label{lem:cpwa_qp}
If $G$ is positive definite, the minimizer is unique and $\Phi$ is a continuous piecewise affine map from $\R^K$ to $\Delta^{K-1}$.
\end{lemma}

$G$ is positive definite if and only if $A$ has full row rank, which requires $K\le d$.

\begin{theorem}\label{thm:relu_sparsemax}
If $G$ is positive definite, there is a finite ReLU network $f_\theta$ such that $\Phi(h)=\sparsemax(f_\theta(h))$ for all $h\in\R^K$.
\end{theorem}

Proofs are in Appendix~\ref{app:proofs}. The theorem establishes exact representability for a fixed polytope with sparsemax; we use softmax (Table~\ref{tab:sparsemax}). For CNNs and AEs, we refit the amortizer every ten epochs and apply $K$ SMO rounds per forward pass. Appendix~\ref{app:amortizerfidelity} evaluates projection fidelity.

\subsection{Token-level PNN.} For pretrained vision backbones we place the polytopal layer on the spatial tokens of the last feature map. Each token $\bz_{n,t}$, $t=1,\dots,T$, of image $n$ is layer-normalized and linearly projected to $256$ dimensions, and all tokens share one polytope $\bA$. The amortizer assigns each token its own coordinates $\bs_{n,t}\in\Delta^{K-1}$, and a linear classifier acts on the token mean of the projections,
\begin{equation}\label{eq:token_pnn}
\mathbf{y}_n=\Big(\tfrac1T\textstyle\sum_{t}\bs_{n,t}\bA\Big)\mathbf{W}^\top+\mathbf{b} .
\end{equation}
The pooled representation therefore stays in the polytope, with coordinates $\bar\bs_n=\tfrac1T\sum_t\bs_{n,t}$. Because the head is affine, the logits equal the mean of the per-token logits $\bs_{n,t}\bA\mathbf{W}^\top+\mathbf{b}$, so every token makes an exact, additive contribution to the prediction. Correspondingly, the corpus consists of individual (image, token) feature rows.

\subsection{Generalizations of PNN}\label{sec:generalization}
Interestingly, the developed PNN methodology naturally reduces to alternative well-known modeling procedures by changing the constraints used in \eqref{eq:PolytopalProjection}.

\begin{figure*}
    \centering
    \includegraphics[width=\linewidth]{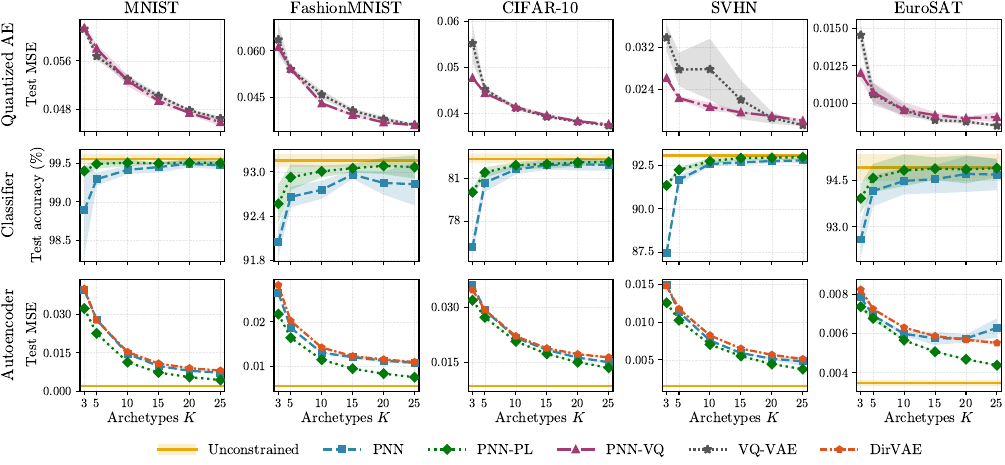}
      \caption{
  Performance as a function of the number of components across datasets. Shaded regions indicate $\pm$ one standard deviation over runs.
  \textbf{Top row:} VQ reconstruction MSE, with improved reconstruction quality reflected by lower values.
  \textbf{Middle row:} Classification accuracy.
  \textbf{Bottom row:} Autoencoder reconstruction MSE.
  }
    \label{fig:combined_performance_vs_k}
\end{figure*}

\textbf{Vector Quantization (VQ): }
Whereas PNNs approximate latent representations via convex combinations of learned archetypes, VQ-VAEs use a finite set of learnable embedding vectors.
Unlike PNNs, where archetypes are learned via gradient-based optimization, the VQ-VAE codebook vectors do not receive gradients directly from the loss. To enable backpropagation through the non-differentiable quantization step, the Straight-Through Estimator (STE) \citep{bengio2013estimating} is employed, passing gradients from the codebook reconstruction $\mathbf{R}^\ell$ to $\mathbf{Z}^\ell$ unchanged. The resulting training objective consists of the task loss and a commitment loss,
\begin{equation}
    \mathcal{L} = \mathcal{L}_{\text{task}} + \beta \sum_{\ell} \|\mathbf{Z}^\ell - \operatorname{sg}[\mathbf{R}^\ell]\|_F^2,
\end{equation}
where $\operatorname{sg}[\cdot]$ is the stop gradient and the commitment term encourages encoder outputs to remain close to their assigned codebook entries and limits latent volume \citep{OordVQVAE}, see also Appendix~\ref{app:AppendixVQ} for further details. 
 Interestingly, the PNN framework provides a direct gradient-based optimization procedure for VQ deep learning by defining $\mathbf{R}^{\ell}$ as the solution to
\begin{equation} \label{eq:VQProjection}
\begin{aligned}
\underset{\mathbf{S}^\ell}{\operatorname{minimize}} \quad & \|\mathbf{Z}^\ell - \mathbf{S}^\ell\mathbf{C}^\ell \mathbf{Z}^{c,\ell} \|_F^2 \\
\text{subject to} \quad & \mathbf{s}^\ell_n \in \{0,1\}^K, \quad \mathbf{s}^\ell_n\mathbf{1} = 1, \quad \forall n.
\end{aligned}
\end{equation}
$\mathbf{s}_n$ is a one-hot vector indicating centroid assignment. Notably, the optimal $\mathbf{C}$ used to define the cluster centroids by minimizing the least squares error is accordingly given by $\mathbf{C}=(\mathbf{S}^{\ell\top}\mathbf{S}^\ell)^{-1}\mathbf{S}^{\ell\top}$, provided every centroid is assigned at least one corpus point. Empirically we found this to always be true, but standard VQ-VAE measures to prevent codebook collapse are also applicable to the PNN-VQ. The equation corresponds to defining the centroids as the average of the observations of the learned corpus $\mathbf{Z}^{c,\ell}$ assigned to the centroid thus satisfying the non-negativity and sum to one constraints making this optimization consistent with the PNN formulation. Consequently, this approach can be directly implemented using the existing constraints on $\mathbf{C}^\ell$ based on the softmax reparameterization, whereas $\mathbf{s}_n$, instead of solved iteratively, can be estimated directly with one non-zero element $p$ by
\begin{eqnarray}
(\mathbf{s}^\ell_n)_p=1 \text{ where } p=\operatorname{argmin}_k \|\mathbf{z}^\ell_n-\mathbf{c}_k\,\mathbf{Z}^{c,\ell}\|_2^2.
\end{eqnarray}
Thereby, the PNN provides a simple gradient-based training objective without the need to impose a commitment loss. The encoder in the PNN-VQ learns through the corpus features $\bZ^{c,\ell}$, and the layers after the projection learn through $\bR^\ell$.

\textbf{Projection Layers (PL):} The use of Projection Layers (PL) in which each layer's activation is projected onto a subspace before further processing is well established within deep learning to perform dimensionality reduction and compression, see also \cite{hawkins2018projection} and the recent generalization to arbitrary higher order feature spaces \cite{morimoto2025tensorprojection}. Such projection layers can naturally be implemented within the PNN framework by imposing unconstrained representations
\begin{equation} \label{eq:Projection}
\begin{array}{rl}
\underset{ \mathbf{S}^\ell}{\operatorname{minimize}} & \|\mathbf{Z}^\ell-\mathbf{S}^\ell\mathbf{C}^\ell \mathbf{Z}^{c,\ell} \|_F^2 \\
\text{subject to} &\quad
 \mathbf{s}^\ell_n\in \mathbb{R}^K \quad \forall n.
\end{array}
\end{equation}  
Consequently, $\mathbf{s}_n$ is trivially given as the closed-form solution to the normal equation
\begin{eqnarray}
\mathbf{s}^\ell_n= \mathbf{z}_n^{\ell}\mathbf{Z}^{c,\ell\top}\mathbf{C^{\ell\top}}\big(\mathbf{C^\ell}\mathbf{Z}^{c,\ell}\mathbf{Z}^{c,\ell\top}\mathbf{C}^{\ell\top}\big)^{-1}.
\end{eqnarray}
whereas $\mathbf{C}^\ell$ is optimized unconstrained without the softmax mapping.  We will denote the above generalizations PNN-VQ and PNN-PL, respectively.

\section{Results and Discussion}\label{sec:results}

\paragraph{Experimental setup.}
We evaluate PNNs on image classification and reconstruction, with additional experiments on tabular classification. The main image benchmarks are MNIST \citep{lecun2010mnist}, FashionMNIST \citep{xiao2017fashionmnistnovelimagedataset}, CIFAR-10 \citep{krizhevsky2009learning}, SVHN \citep{netzer2011svhn}, and EuroSAT \citep{EUROSAT}. These experiments use a CNN classifier with polytopal layers after its two fully connected hidden layers and a convolutional autoencoder (AE) with a polytopal bottleneck. We further evaluate the PNN-CNN on PathMNIST, DermaMNIST, and BloodMNIST from MedMNIST v2 \citep{Yang2023MedMNISTClassification}. For larger backbones and token representations, we evaluate ImageNet-100 \citep{tian2020contrastivemultiviewcoding} and ImageNet-1k \citep{deng2009imagenet} using ConvNeXt-T \citep{liu2022convnet2020s}, initialized from ImageNet-1k weights and fine-tuned on each benchmark. The PNN projects each spatial token onto a shared polytope before averaging the resulting representations and applying a linear classifier. We also evaluate a PNN head on the pooled features of a frozen compact vision transformer \citep{hassani2022escapingbigdataparadigm} on CIFAR-10.
Additional MLP results on Digits \citep{optical_recognition_of_handwritten_digits_80}, Iris \citep{iris_53}, Wine \citep{wine_109}, and Breast Cancer \citep{breast_cancer_wisconsin_diagnostic_17} are reported in Appendix~\ref{app:mlp}.

The main CNN and AE experiments compare PNN with the corresponding unconstrained networks and the PNN-VQ and PNN-PL variants, enabling comparison between polytopal, subspace-based, and codebook-based representations. The AE comparisons additionally include VQ-VAE \citep{OordVQVAE} and a variational autoencoder with a Dirichlet bottleneck \citep{joo2020dirichlet}. The MedMNIST and larger-backbone experiments compare PNNs with their corresponding unconstrained models. Dataset splits, architectures, training procedures, and experiment-specific settings are provided in Appendix~\ref{app:ExperimentalDetails}.

We first evaluate the proposed scalable inference framework by comparing mini-batched training with full-data (non-batched) updates. Mini-batching with batch-wise corpus updates (Section~\ref{subsection:corpus_updates}) achieves accuracy comparable to full-data training (see Appendix \autoref{tab:b_nb_accuracy}). We also evaluate the proposed amortized simplex inference procedure and find that with periodic realignment this serves as a fast and accurate replacement for the full QP projection (Appendix~\ref{app:amortizerfidelity}).

To validate our direct training strategy for PNN-VQ, we compare it to standard VQ training with a commitment loss \citep{OordVQVAE}. As shown in \autoref{fig:combined_performance_vs_k}, PNN-VQ achieves comparable, and in several cases lower test mean-squared-error (MSE), with detailed results reported in \autoref{tab:vq-full}.

We next evaluate classification with a CNN backbone. In \autoref{fig:combined_performance_vs_k}, the PNN generally approaches the accuracy of the unconstrained CNN as the number of archetypes increases. The encoder dimension remains fixed, while larger values of $K$ allow the constrained representation to use more archetypes. The largest improvements occur when moving
away from the smallest archetype counts, although the remaining gap to the unconstrained model varies across datasets.

This comparison extends to the three MedMNIST datasets. With $K=25$, the PNN achieves test accuracies of 89.1\% on PathMNIST, 77.4\% on DermaMNIST, and 98.1\% on BloodMNIST, compared with 90.4\%, 78.6\%, and 98.3\% for the corresponding unconstrained CNNs (\autoref{tab:medmnist}). Increasing $K$ from 10 to 25 improves the mean accuracy on PathMNIST and DermaMNIST, while BloodMNIST remains at 98.1\%.

We evaluate the proposed method in an unsupervised setting using an AE. Reconstruction quality is measured using MSE, with a single PNN layer structuring the bottleneck under the PNN constraints. As shown in \autoref{fig:combined_performance_vs_k}, the PNN-AE shows similar tendencies as the PNN-CNN and generally performs competitively to the less constrained variants. PNN-AE generally achieves lower reconstruction MSE than DirVAE. The unconstrained AE and PNN-PL generally achieve lower MSE than PNN-AE, indicating a reconstruction cost associated with the convex constraint that generally shrinks with $K$. We also compare softmax and sparsemax as the amortizer output map (\autoref{tab:sparsemax}). Across MNIST, FashionMNIST, and CIFAR-10, the two variants achieve similar reconstruction MSE under the same refinement protocol.

Token-level PNNs on pretrained ConvNeXt-T retain most of the unconstrained model’s accuracy (\autoref{tab:scale}). On ImageNet-100, accuracy varies little for $K\ge100$, with overlapping error bars. Full results are in Appendix~\ref{app:scale}.

Next, the PNN module is added to a frozen compact transformer where PNN acts on the pooled representation.
On CIFAR-10, its PNN head achieves 94.46\% test accuracy with $K=10$, compared to 94.47\% for the unconstrained
classifier (\autoref{tab:scale}). 

\begin{table}[t]
\caption{Larger backbones and label spaces. Accuracy (\%) of the unconstrained model and of the PNN, and the share of unconstrained accuracy the PNN retains. IN-100 and IN-1k are ImageNet-100 and ImageNet-1k, whose backbones are fine-tuned. Full results in Appendix~\ref{app:scale}.}\label{tab:scale}
\begin{center}\scriptsize
\setlength{\tabcolsep}{2pt}
\resizebox{\linewidth}{!}{\begin{tabular}{lccccc}
\toprule
Setting & Classes & Unconstr. & $K$ & PNN & Kept (\%) \\
\midrule
IN-100, ConvNeXt-T & 100 & 94.6$\pm$0.1 & 25 & 91.7$\pm$0.5 & 96.9 \\
IN-1k, ConvNeXt-T & 1000 & 81.2$\pm$0.0 & 128 & 78.6$\pm$0.2 & 96.9 \\
CIFAR-10, compact ViT & 10 & 94.47$\pm$0.21 & 10 & 94.46$\pm$0.23 & 100.0 \\
\bottomrule
\end{tabular}}
\end{center}
\end{table}

\begin{figure}
    \centering
    \includegraphics[width=\linewidth]{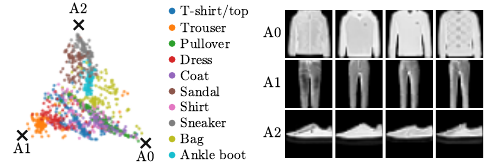}
    \caption{PNN-AE ($K=3$) on FashionMNIST. Left: latent space projected via MDS; crosses indicate archetypes. Right: the four training images with the largest weight on each archetype, showing coherent groups (upper body, lower body, footwear).}
    \label{fig:qualitativeK3FMNIST}
\end{figure}

\autoref{fig:qualitativeK3FMNIST} shows the latent space of a PNN-AE with $K=3$ on FashionMNIST using multidimensional scaling, together with the four training images with the largest weight on each archetype. The displayed examples form recognizable groups of lower-body clothing, upper-body clothing, and footwear. This gives a qualitative illustration of how the learned polytope organizes the representations in this model.

\begin{figure}
    \centering
    \includegraphics[width=\linewidth]{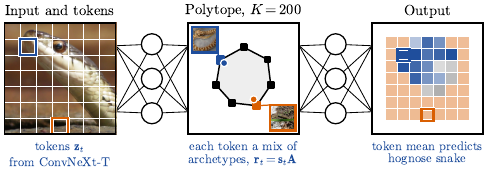}
    \caption{Token-level PNN on ImageNet-100 with ConvNeXt-T and 200 archetypes. Each spatial token is represented as a convex combination of shared archetypes. A linear classifier reads the mean of the reconstructed token features. Two archetypes associated with the predicted and competing classes are illustrated using training images. The outlined tokens have the largest weights on the displayed archetypes.}
    \label{fig:pnn_token_architecture}
\end{figure}

\begin{figure}[ht]
    \centering
    \includegraphics[width=\linewidth]{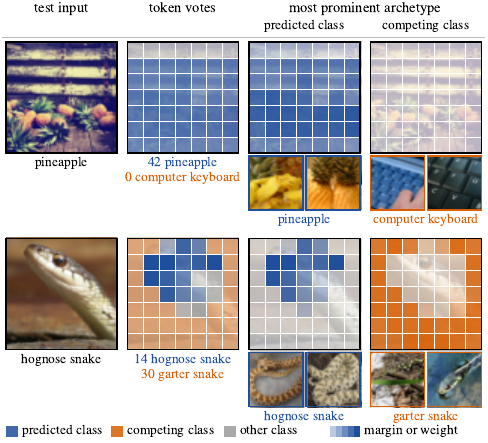}
    \caption{Token predictions and archetype weights for two ImageNet-100 images. Each row shows the input, token predictions, and spatial weights of selected archetypes associated with the predicted and competing classes. Training examples are used to illustrate the archetypes. Blue denotes the predicted class, orange the competing class, and gray other classes. The counts below the token-prediction panels show how many tokens favor each displayed class. The image prediction depends on the averaged class scores.}
    \label{fig:pnn_token_Example}
\end{figure}

The token-based ImageNet models allow the same archetypal representation to be inspected spatially.
\autoref{fig:pnn_token_architecture} illustrates how individual tokens are represented using a shared polytope before their reconstructed features are averaged for classification. \autoref{fig:pnn_token_Example} shows token predictions and the spatial weights of selected archetypes for individual images. The associated training examples provide visual references for interpreting these archetypes.

Each token is represented by its coordinates inside the polytope. The image-level margin can therefore be described as a sum of archetype scores, each weighted by the archetype's share of the image (Figure \ref{fig:pnn_token_Example}). The snake example is described by two archetypes. Most of the tokens fall near an archetype whose training images show a small snake within a scene. The head tokens are primarily described by an archetype whose training images shows the hognose snake scale pattern. From the token votes we see that the head tokens have higher weight which influences the predictions. Unlike prototype networks \citep{chen2019looks,Nauta_2023_CVPR}, which score patches by similarity to learned prototypes, the PNN represents each token as a point in the polytope spanned by the archetypes, so the decomposition uses the token's own coordinates and requires no prototype-specific losses.


From a computational perspective, PNNs add per-iteration cost through corpus-based archetype construction and simplex inference. This cost depends on the batch size, representation dimension, corpus size, number of archetypes, and number of tokens. The inference procedure differs across experiments. The CNN and AE use the amortized simplex infernece procedure followed by $K$ SMO refinement rounds during training and evaluation. The ImageNet models use the amortizer without this refinement in the task path. The larger-scale experiments demonstrate that the construction can be used with spatial token representations, while its computational overhead depends on the chosen inference procedure. Costs are further examined in Appendix~\ref{app:costs}.
From an explainability perspective, PNNs expose the archetypal decomposition of a representation used in the forward computation. The FashionMNIST and ImageNet examples show how this decomposition can be inspected at the level of an entire input or individual spatial tokens. The archetypes are learned without predefined concept annotations, and representative training examples provide a way to examine their visual meaning. For the ImageNet model's linear classifier, the archetype contributions can also be traced directly to the class scores. With nonlinear downstream networks, the coordinates describe the intermediate representation, and their effect on the output must be assessed through that subsequent computation. We test this by removing the archetype with the largest simplex coordinate, renormalizing the remaining coordinates, and compared this with removing a random archetype. For the CNN, this changes more predictions than removing a random archetype (Figures~\ref{fig:aedeletionbyk} and \ref{fig:cnndeletionbyk}).

\section{Conclusion}
We have introduced the Polytopal Neural Network, a novel architecture that prioritizes interpretability and trustworthiness in deep learning by rigorously enforcing AA constrained latent representations . Through extensive evaluation across multiple network architectures and diverse datasets, we have demonstrated that PNNs produce inherently interpretable predictions with semantically meaningful explanations. Unlike post-hoc explanation methods, PNNs provide a principled guarantee that the explanations reflect the actual decision-making process learned during training, addressing a fundamental challenge in XAI.
Our analysis reveals that PNNs can be understood as a continuous quantization method, establishing a natural connection to vector quantization approaches. In this framework, the learned archetypes form a semantically meaningful codebook that enables both effective classification and transparent reasoning. 

\bibliographystyle{plainnat}  
\bibliography{example_paper}       

\clearpage
\appendix
\thispagestyle{empty}

\onecolumn
\aistatstitle{The Polytopal Neural Network \\
Supplementary Materials}
\section{Proofs}\label{app:proofs}

We use the notation of Section~\ref{sec:theory}. The archetype matrix $A\in\R^{K\times d}$ is fixed, $G=AA^{\top}$, and $\Delta^{K-1}=\{s\in\R^K\mid s\ge0,\ \one^\top s=1\}$.
Write
\begin{equation}
    F_h(s)=\tfrac12 s^\top Gs-h^\top s.
\end{equation}

\subsection{Proof of Lemma~\ref{lem:cpwa_qp}}

Since $G$ is positive definite, $F_h$ is strictly convex. The simplex is compact and convex, so a minimizer exists and is unique for every $h\in\R^K$.

Fix a nonempty index set $\mathcal J\subseteq\{1,\ldots,K\}$ and set $s_{\mathcal J^c}=0$. The stationarity and sum-to-one conditions on these coordinates are
\begin{equation}
    \begin{pmatrix}
G_{\mathcal J\mathcal J} & \one\\
\one^\top & 0
\end{pmatrix}
\begin{pmatrix}
s_{\mathcal J}\\
\lambda
\end{pmatrix}
=
\begin{pmatrix}
h_{\mathcal J}\\
1
\end{pmatrix}.
\end{equation}
Here $\lambda$ is the multiplier for $\one^\top s=1$. Because $G_{\mathcal J\mathcal J}$ is positive definite, this system is invertible. Its solution $s^{\mathcal J}(h)$ and $\lambda^{\mathcal J}(h)$ is therefore affine in $h$.

This candidate satisfies the remaining KKT conditions exactly when
\begin{equation}
    s^{\mathcal J}_{\mathcal J}(h)\ge0,
\qquad
\bigl(
G s^{\mathcal J}(h)-h+
\lambda^{\mathcal J}(h)\one
\bigr)_{\mathcal J^c}\ge0.
\end{equation}
These inequalities are affine in $h$ and define a closed polyhedral region. On this region the candidate is the unique minimizer, so $\Phi$ is affine there.

Every minimizer has a nonempty support and satisfies the KKT conditions for that support. The regions therefore cover $\R^K$. There are finitely many index sets, and uniqueness ensures that the affine formulas agree wherever their regions overlap.

Finally, let $h_m\to h$. Compactness of the simplex ensures that every subsequence of $\Phi(h_m)$ has a convergent further subsequence. Passing to the limit in
\begin{equation}
    F_{h_m}(\Phi(h_m))\le F_{h_m}(s)
\qquad\text{for every }s\in\Delta^{K-1}
\end{equation}
shows that every such limit minimizes $F_h$. Uniqueness forces the limit to equal $\Phi(h)$. Thus $\Phi(h_m)\to\Phi(h)$, establishing continuity. Hence $\Phi$ is continuous piecewise affine. \qed

\subsection{Proof of Theorem~\ref{thm:relu_sparsemax}}

By Lemma~\ref{lem:cpwa_qp}, each coordinate of $\Phi$ is a continuous piecewise affine function with finitely many pieces. Every such scalar function admits an exact finite ReLU representation \citep{arora16}. Representing each coordinate and stacking the resulting networks gives a finite ReLU network $f_\theta$ satisfying
\begin{equation}
    f_\theta(h)=\Phi(h)
\qquad\text{for all }h\in\R^K.
\end{equation}
Sparsemax is the Euclidean projection onto the probability simplex \citep{Martins2016},
\begin{equation}
    \sparsemax(v)=
\argmin_{s\in\Delta^{K-1}}\tfrac12\|s-v\|^2.
\end{equation}
It therefore leaves every point of $\Delta^{K-1}$ unchanged. Since $\Phi(h)\in\Delta^{K-1}$,
\begin{equation}
    \sparsemax(f_\theta(h))
=\sparsemax(\Phi(h))
=\Phi(h)
\qquad\text{for all }h\in\R^K.
\end{equation}
This proves the theorem. \qed

\section{On the simplex projection}\label{app:S}

\subsection{Quadratic Scaling of Projections via. SMO-Updates}\label{app:smo}

Instead of the QP based estimation producing the full gradient of $\mathbf{C}^\ell$ accounting also for the $\mathbf{S}^\ell$ dependencies on $\mathbf{C}^\ell$ as given by \eqref{eq:CdependencyS}:

\begin{equation}\label{eq:CdependencyS}
\mathbf{r}_n^\top=(\mathbf{C^\ell}\mathbf{z}_n^\ell)_\mathcal{P}^\top (\mathbf{C^\ell}\mathbf{Z}^\ell\mathbf{z}_n^{\ell\top}+\lambda\boldsymbol{1})_\mathcal{P}^\top \big(\mathbf{C^\ell}\mathbf{Z}^\ell\mathbf{Z}^{\ell\top}\mathbf{C}^{\ell\top}+\lambda\mathbf{I}\big)_{\mathcal{P},\mathcal{P}}^{-1},
\end{equation}

we can reduce the gradient computation of $\mathbf{C}^\ell$ to only account for the direct influence that $\mathbf{C}^\ell$ exerts on $\mathbf{r}_n$ keeping $\mathbf{s}_n$ fixed. Consequently, we propagate gradients only through the expression
$\mathbf{r}_n^\ell =\mathbf{s}^\ell_n \mathbf{C}^\ell \mathbf{Z}^\ell$,
treating $\mathbf{s}^\ell_n$ as constant. This avoids having gradient computations rely on any cubic in scaling matrix inversion.

To further reduce the cubic complexity of estimating $\mathbf{s}_n$ we exploit the recently proposed 
sequential minimal optimization (SMO) procedure for AA \cite{AABinary}. The SMO procedure only considers two coefficients $(p,q)$ at a time. Defining
\begin{align}
\sigma_i^\ell= s^\ell_{i,p} + s^\ell_{i,q},\qquad  
s^\ell_{i,p}=\sigma^\ell_i\alpha^\ell_i, \qquad 
s^\ell_{i,q}=\sigma^\ell_i(1-\alpha_i^\ell),
\end{align}
the problem reduces to a one-dimensional quadratic function in $\alpha_i^\ell \in [0,1]$ redistributing the mass $\sigma_i^\ell$ between the two elements. Importantly, this has a simple closed-form solution given by
\begin{equation}
\label{eq:alpha_update}
\alpha_i^{\star\ell}
=
\underset{{[0,1]}}{\mathrm{clip}}
\left(
-
\frac{
\sigma_i (H^\ell_{qp}-H^\ell_{qq}) + d^\ell_{i,p}-d^\ell_{i,q}
}{
\sigma_i (H^\ell_{pp}-2H^\ell_{pq}+H^\ell_{qq})
}
\right),
\end{equation}
where $\mathbf{H}^\ell=(\mathbf{C}^\ell\mathbf{Z}^\ell)(\mathbf{C}^\ell\mathbf{Z}^\ell)^\top$,
$\mathbf{d}^\ell_i=-\mathbf{z}^\ell_i(\mathbf{C}^\ell\mathbf{Z}^\ell)^\top+\mathbf{s}^\ell_i\mathbf{H}^\ell$.

Pairs with $\sigma_i=0$ are skipped, as they contain no weight. 

Pairs $(p,q)$ are iterated in random order, optionally multiple times per batch in which it was found in \cite{AABinary} that convergence could be achieved with complexity $\mathcal{O}(K^2)$. All updates are performed outside the computational graph such that gradients are propagated only through archetype parameters $\mathbf{C}^{\ell}$, not through $\mathbf{S}$ and can be implemented trivially in parallel across the considered observations. This yields substantial speedups while preserving high-quality simplex-constrained projections.

\subsection{Fidelity of the Amortizer}\label{app:amortizerfidelity}
\begin{figure}
    \centering
    \includegraphics[width=0.5\linewidth]{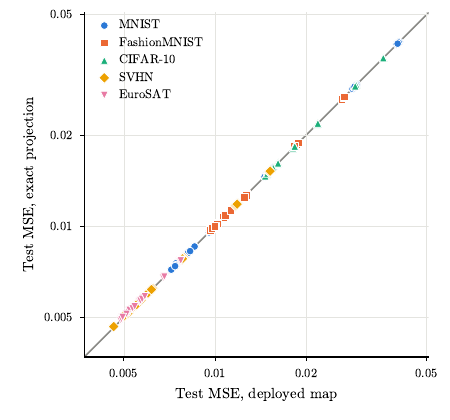}
    \caption{Test MSE of the PNN-AE with the exact SMO projections vs. the Amortized version.}
    \label{fig:amortizerFidelity}
\end{figure}

To test the fidelity of the Amortizer  we replace the deployed amortizer with the exact SMO projection at test time. Figure~\ref{fig:amortizerFidelity} shows the resulting test MSE against the amortized test MSE. All 150 configurations lie on the diagonal. The amortizer therefore does not materially change what the trained network computes.

\section{Interpretability of the Polytope}\label{app:interpretability}

\begin{figure}[ht]
    \centering
    \includegraphics[width=\linewidth]{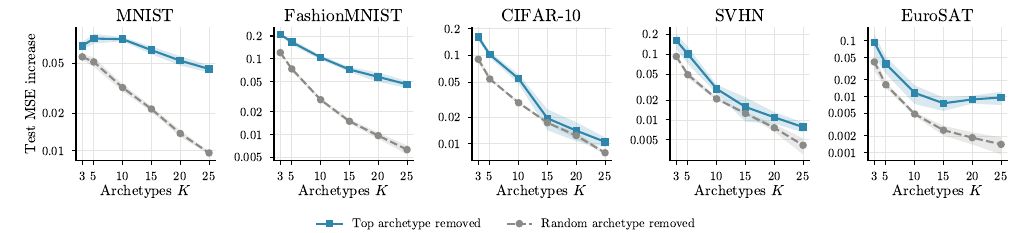}
    \caption{Share of PNN-CNN test predictions that change when the highest-weight archetype is removed and when a single archetype is removed at random, against the number of archetypes $K$. Mean $\pm$ standard deviation for five seeds.}
    \label{fig:aedeletionbyk}
\end{figure}

\begin{figure}[ht]
    \centering
    \includegraphics[width=\linewidth]{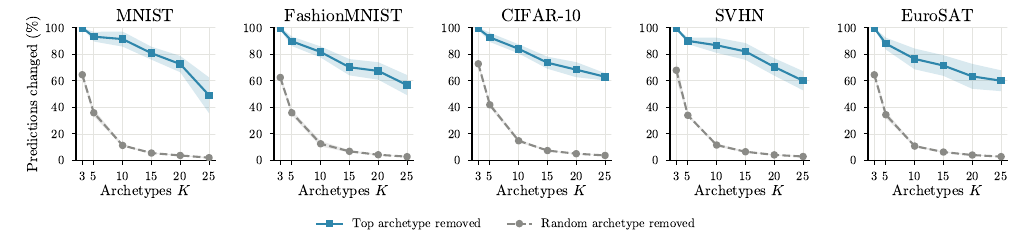}
    \caption{Increase in test MSE of the PNN-AE when the highest-weight archetype is removed and when a single archetype is removed at random, against the number of archetypes $K$. Mean $\pm$ standard deviation for five seeds.}
    \label{fig:cnndeletionbyk}
\end{figure}

To test the interpretability of the PNN, we test how the output depends on the archetypes, if the archetypes are useful extreme configurations, removing the top archetype should hurt performance more than removing archetypes at random. We therefore take every trained PNN-AE and remove one archetype from the deployed coordinates of 1000 test images. The remaining coordinates are renormalized and decoded by the same decoder. Figures~\ref{fig:aedeletionbyk} and \ref{fig:cnndeletionbyk} shows both increases against $K$. 

\section{Token-wise Representations}\label{app:tokens}
To make the method applicable to larger scale models with larger backbones we implement a token based version of the PNN. It works by applying the projection along the feature dimension and applying it to each token of a sequence. A patch embedding produces $\bZ = [\bz_1,\dots,\bz_T]\in\R^{T\times d}$, and each token is projected onto the same polytope, $\br_t = \bs_t\bA$ with $\bA=\bC\bZ^{c}$. Stacking gives $\bR = \bS\bA$ with $\bS\in\R^{T\times K}$. In ImageNet, each corpus element is the cached feature vector of one training image and token. These vectors are recomputed from the current backbone every epoch, and column generation updates corpus membership. The same corpus is shared across positions.

\section{Experimental Details}\label{app:ExperimentalDetails}
\subsection{Data} \label{app:data}
MNIST, FashionMNIST, CIFAR-10 and SVHN use their official test splits. We hold out 10\% of each training split for validation. EuroSAT uses a 70/10/20 training, validation and test split. The CNN splits are stratified by class, while the AE splits are unstratified. Each run uses its training seed for the split. MNIST and FashionMNIST images are $28\times28$, CIFAR-10 and SVHN images are $32\times32$, and EuroSAT images are $64\times64$.

Classifier inputs use fixed channel normalization. CIFAR-10, SVHN and EuroSAT training batches use reflected random crops with padding 4 and horizontal flips with probability 0.5. MNIST and FashionMNIST use reflected random crops with padding 2 and no flips.

\paragraph{Other experiments.} PathMNIST, DermaMNIST and BloodMNIST from MedMNIST v2 \citep{Yang2023MedMNISTClassification} use their official splits at $64\times64$ and the classifier augmentation above. ImageNet-100 \citep{tian2020contrastivemultiviewcoding} is the class subset of ImageNet-1k \citep{deng2009imagenet}. Both train on random resized crops with flips and evaluate on a centre crop. The frozen transformer uses CIFAR-10 with 5000 training images held out for validation.

\subsection{Architectures} \label{app:archetectures}

The image encoder has three strided convolutions, $\mathrm{Conv}_{3\times3,s=2}(C,32)$, $\mathrm{Conv}_{3\times3,s=2}(32,64)$ and $\mathrm{Conv}_{3\times3,s=2}(64,128)$, each followed by group normalization with 8, 16 and 32 groups and a ReLU. The feature map is flattened and linearly projected to $\bz\in\R^{64}$. The classifier applies $\bh_1=\mathrm{ReLU}(W_1\bz)$, a PNN projection $\tilde\bh_1$, $\bh_2=\mathrm{ReLU}(W_2\tilde\bh_1)$, a second PNN projection $\tilde\bh_2$, and a linear output layer. The first site forms archetypes from their encoded features. Before the second site, the corpus passes through the first PNN projection and the second fully connected layer, as the input images do. Each site forms its archetypes from the corpus features entering that site. The autoencoders use the same encoder, a PNN bottleneck, and a mirrored transposed-convolution decoder with group normalization, ReLU and a sigmoid output.

\paragraph{Other experiments.} The MedMNIST classifier is the CNN above with a 128-dimensional latent space. On ImageNet, ConvNeXt-T start from ImageNet-1k weights. Their spatial tokens are projected to 256 dimensions by a PCA-initialized linear layer, and the PNN projects every token (Appendix~\ref{app:tokens}). The reconstructions are averaged over tokens before a linear classifier. The unconstrained model applies the same projection to the pooled feature, and the width control uses 128 dimensions. The frozen-head experiment places the PNN after the attention pooling of a compact transformer \citep{hassani2022escapingbigdataparadigm} and initializes its output layer from the backbone classifier.

\subsection{Model Variants} \label{app:modelvariants}
\paragraph{PNN.} The amortizer of Eq.~\eqref{eq:amortizer} has two hidden layers of width $\max(8K,128)$ and ReLU activations. The CNN and AE refine its detached output with $K$ SMO rounds during both training and evaluation. 
\paragraph{PNN-VQ.} One-hot coordinates by nearest archetype, with the reconstruction or task loss only. In the autoencoder, $\bC$ is initialized from $k$-means clusters of the encoded corpus, using a softened form of $(\bS^\top\bS)^{-1}\bS^\top$ from Section~\ref{sec:generalization} with 95\% of each row of $\bC$ spread uniformly over its cluster and the remaining 5\% spread uniformly over the other corpus points, and is then learned by the task gradient.
\paragraph{PNN-PL.} Unconstrained $\bC$ and least squares coordinates. 
\paragraph{VQ-VAE.}\label{app:vq} A codebook updated by an exponential moving average, the straight-through estimator, and the commitment loss $\beta\|\bZ-\mathrm{sg}[\bR]\|_F^2$ with $\beta=0.25$ \citep{OordVQVAE}. The codebook is initialized by $k$-means++ and unused entries are reinitialized. 
\paragraph{DirVAE.} The encoder outputs positive concentrations. Training uses the inverse Gamma CDF approximation and normalizes the samples onto the simplex \citep{joo2020dirichlet}. Evaluation uses the posterior mean. The loss adds $\beta_{\mathrm{KL}}D^{-1}\sum_k\mathrm{KL}(\mathrm{Gamma}(\alpha_k,1)\,\|\,\mathrm{Gamma}(1,1))$ to mean pixel MSE, where $D$ counts pixel values and $\beta_{\mathrm{KL}}=0.01$. 

\subsection{Training Protocol} \label{app:trainingprotocol}

The main CNN and AE grids use $K\in\{3,5,10,15,20,25\}$ and seeds $\{42,123,456,789,1024\}$. Runs use AdamW, cosine learning-rate schedules and gradient-norm clipping. Table~\ref{tab:hparams} lists epoch limits and initial learning rates. Early stopping can shorten each phase.

Phase 1 loads the unconstrained network trained with the same seed. Before Phase 2, AA on up to 20,000 encoded training images selects one anchor image per corpus point. The anchor starts with weight 0.25 and the remaining weight is spread uniformly over the training set. The archetypes are initialized on this corpus. Phase 2 trains the archetype weights and output layer, with the encoder frozen. The CNN also freezes its two inner fully connected layers. Phase 3 updates the task network after an initial period with the encoder frozen. The corpus logits are trained by the task loss through the mini-batch rows of Eq.~\eqref{eq:streaming_XC}.

\paragraph{Amortizer fitting.} The AE fits the amortizer by minimizing $\|z-\hat s A\|^2$, normalized by the mean squared feature magnitude. CNN warm starts and refits minimize coordinate MSE against targets computed with $2K$ batched SMO rounds. Features, archetypes and targets are detached during fitting. Between refits, each training step also takes one amortizer step on the normalized projection residual $|\bz-\hat\bs\bA|^2/\overline{|\bz|^2}$, with $\bz$ and $\bA$ detached. This step updates only the amortizer, and the task loss never updates it.

Refits run every ten epochs and after Phase 2. The AE also refits after Phase 3. Its final refit is retained only if the measured projection residual on training data does not increase. The CNN has no post-Phase-3 refit.

\begin{table}[ht]
\caption{Hyperparameters of the main CNN and AE grids. Epoch counts are maxima and learning rates are initial values.}\label{tab:hparams}
\begin{center}\footnotesize
\begin{tabular}{lcc}
\toprule
 & CNN classifier & Autoencoder \\
\midrule
Latent dimension & 64 & 64 \\
Corpus size & 2048 & 2048 \\
Baseline batch size & 256 & 2048 \\
Phase 2 and Phase 3 batch size & 512 & 512 \\
Unconstrained baseline, epochs & 200 & 200 \\
Phase 2 epochs & 60 & 100 \\
Phase 3 epochs & 300 & 200 \\
Backbone frozen at the start of Phase 3 & 30 epochs & 20 epochs \\
Learning rate, unconstrained baseline & $10^{-3}$ & $10^{-3}$ \\
Learning rate, archetypes in Phase 2 & $3\cdot10^{-3}$ & $5\cdot10^{-3}$ \\
Learning rate, output layer in Phase 2 & $5\cdot10^{-4}$ & $2\cdot10^{-3}$ \\
Learning rate in Phase 3, encoder / output / archetypes & $10^{-4}$ / $2\cdot10^{-4}$ / $10^{-3}$ & $10^{-4}$ / $2\cdot10^{-4}$ / $10^{-3}$ \\
Learning rate, CNN inner layers in Phase 3 & $10^{-4}$ & -- \\
Gradient clipping & 2.0 & 1.0 \\
Amortizer epochs, warm start / refit after Phase 2 / after Phase 3 & 200 / 100 / 0 & 200 / 100 / 100 \\
Amortizer refit interval / maximum epochs per refit & 10 / 30 & 10 / 30 \\
Initial amortizer temperature $\tau$ & 1 & 1 \\
Learning rate, corpus logits, Phase 2 / Phase 3 & $2\cdot10^{-2}$ / $5\cdot10^{-3}$ & $10^{-2}$ / $10^{-3}$ \\

\bottomrule
\end{tabular}
\end{center}
\end{table}

\paragraph{Task objective and adaptation.} The task loss is pixel MSE for the AE and cross-entropy for the CNN, with no latent AA penalty. After periodic amortizer refits, two epochs adapt the output layer at learning rate $10^{-3}$. The AE also allows up to 50 decoder-only adaptation epochs before Phase 3. The layer-count and simplex-map ablations use seeds $\{42,123,456\}$ and $K\in\{3,10,25\}$. Their comparison arms use the same seeds.

\subsection{Training beyond the main grids}\label{app:details-other}

Table~\ref{tab:hparams-other} summarizes the MedMNIST, ImageNet and frozen-transformer experiments. Appendices~\ref{app:data} and \ref{app:archetectures} describe their data and architectures. They share the PNN layer, AA corpus selection, column generation and an amortizer that the task loss does not update. They differ in backbone, optimizer, coordinate map and reported metric.

\begin{table}[!ht]
\caption{Settings of the experiments beyond the main CNN and AE grids. Unconstrained models use the same backbone, data and optimization as the PNN of their column. The ImageNet-100 PNN has three seeds and its unconstrained model five.}\label{tab:hparams-other}
\begin{center}\footnotesize
\setlength{\tabcolsep}{4pt}
\begin{tabular}{lcccc}
\toprule
 & MedMNIST & ImageNet-100 & ImageNet-1k & Frozen head \\
\midrule
Backbone & two-site CNN & ConvNeXt-T & ConvNeXt-T & compact ViT \\
Backbone initialization & baseline, same seed & ImageNet-1k & ImageNet-1k & CIFAR-10, same seed \\
Input resolution & $64\times64$ & $224\times224$ & $224\times224$ & $32\times32$ \\
PNN site & two FC sites & every token & every token & pooled feature \\
PNN input dimension & 128 & 256 & 256 & 256 \\
$K$ & 10, 25 & 25, 50, 100, 150, 200 & 128, 256 & 3, 5, 10, 20, 40 \\
Corpus & 2048 images & 3200 token rows & 32,768 token rows & 2048 feature rows \\
Seeds & 5 & 3 & 2 & 5 \\
Optimizer & AdamW & SGD & SGD & AdamW \\
Batch size & 256 & 512 & 512 & 512 \\
Epochs & 80 + 160 & 2 + 30 & 2 + 30 & 80 \\
Amortizer hidden width & $\max(8K,128)$ & 1024 & 1024 & $\max(8K,128)$ \\
Coordinate map & $K$ refinement rounds & amortizer only & amortizer only & $K$ refinement rounds \\
Reported split & test & validation & validation & test \\
Checkpoint & best validation & Final epoch & Final epoch & best validation \\
\bottomrule
\end{tabular}
\end{center}
\end{table}

\paragraph{MedMNIST.} Training follows the main CNN protocol with the changes in Table~\ref{tab:hparams-other}. Phase 2 and Phase 3 run for up to 80 and 160 epochs, the encoder stays frozen for the first 20 Phase 3 epochs, and the post-Phase-2 amortizer refit is limited to 50 epochs.

\paragraph{ImageNet.} Two epochs with the backbone frozen are followed by 30 epochs of fine-tuning with SGD, learning rate 0.02, a backbone multiplier of 0.1, momentum 0.9, weight decay $10^{-4}$, label smoothing 0.1 and cosine decay. As in the main grids, AA selects the corpus from training tokens and column generation updates it. The amortizer is fitted to projected-gradient targets and refitted when its arg-max agreement falls below 0.85. It is deployed without SMO refinement.

\paragraph{Frozen transformer head.} Each backbone is trained for 160 epochs with mixup, CutMix, RandAugment and random erasing. The head trains for 80 epochs with AdamW, learning rate $10^{-3}$ and cosine decay, using $K$ refinement rounds. Its amortizer is refitted to $2K$-round SMO targets every ten epochs, when column generation also updates the corpus.

The released code records every remaining setting.

\subsection{Choice of K} \label{app:K}
We define $K$ as the number of archetypes that span the convex polytope approximating the latent representation manifold. This hyperparameter controls the geometric expressivity of the model, meaning that smaller values of $K$ impose a stronger convex constraint, leading to a more compact and interpretable representation, whereas larger values increase the capacity to capture fine-grained variations in the data. Thus, $K$ governs a trade-off between compression and fidelity. From a geometric perspective, increasing $K$ refines the polytope approximation of the underlying data manifold, enabling the network to capture higher-order structures while maintaining interpretability through the archetypal basis. We recommend systematically assessing the impact of the choice of $K$ on performance as assessed in \autoref{fig:combined_performance_vs_k} and for ease of interpretation select the smallest $K$ with sufficient performance. We presently used the same number of archetypes for each PNN layer. Whereas this can be defined for each layer separately, tuning separate values of $K$ for each layer requires the fitting of many models and future work should investigate efficient search strategies for identifying suitable layer specific values of $K$.

\subsection{Training costs} \label{app:costs}

\begin{figure*}[t]
\centering
\includegraphics[width=\textwidth]{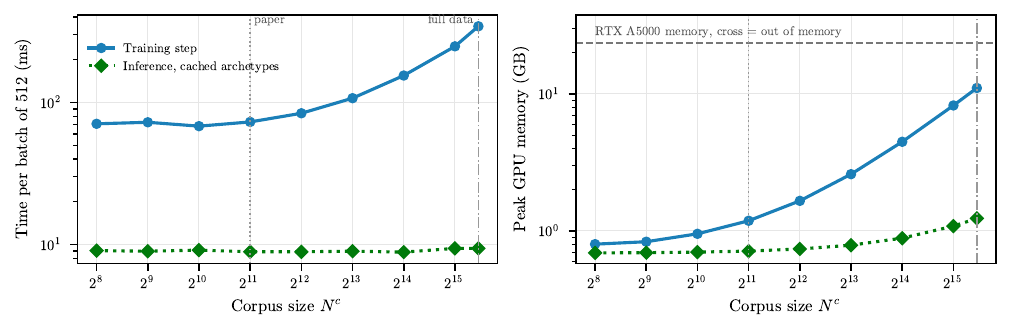}
\caption{Cost of the PNN-CNN (CIFAR-10, $K=10$) against the corpus size $N^c$, on one NVIDIA RTX A5000. Left: time per batch of 512 for a Phase-3 training step and for inference with the archetypes cached. Right: peak GPU memory of the same. Using the full training set as the corpus ($N^c=45,000$) costs 4.7$\times$ the training time and 9.3$\times$ the memory of the learned corpus ($N^c=2048$); inference time does not depend on $N^c$.}\label{fig:corpus-cost}
\end{figure*}

Table~\ref{tab:cnn-cost} shows that training time increases with $K$, while memory changes only slightly. As expected PNN-PL and PNN-VQ are cheaper than PNN at the same $K$.
Figure~\ref{fig:corpus-cost} shows that a smaller corpus reduces training time and memory.

\begin{table*}[ht]
\caption{Measured training cost of the CNN models on one NVIDIA RTX A5000. Each model trains for 100 epochs from its seed-42 checkpoint with its own training step.}\label{tab:cnn-cost}
\begin{center}\small
\setlength{\tabcolsep}{4.8pt}
\begin{tabular}{lcccccc}
\toprule
 & Unconstrained & PNN $K=3$ & PNN $K=10$ & PNN $K=25$ & PNN-PL $K=25$ & PNN-VQ $K=25$ \\
\midrule
\multicolumn{7}{l}{\emph{Training time (minutes per 100 epochs)}} \\
MNIST & 5.3 & 10.3 & 14.9 & 24.8 & 7.7 & 6.7 \\
FMNIST & 5.3 & 10.3 & 14.9 & 25.5 & 7.8 & 6.6 \\
CIFAR-10 & 4.5 & 9.0 & 12.8 & 21.1 & 6.6 & 5.9 \\
SVHN & 6.5 & 14.2 & 18.8 & 31.4 & 9.7 & 8.5 \\
EuroSAT & 2.1 & 5.4 & 7.0 & 10.5 & 4.2 & 3.7 \\
\midrule
\multicolumn{7}{l}{\emph{Peak GPU memory in training (GB)}} \\
MNIST & 0.24 & 0.60 & 0.61 & 0.64 & 0.52 & 0.52 \\
FMNIST & 0.24 & 0.60 & 0.61 & 0.64 & 0.52 & 0.52 \\
CIFAR-10 & 0.70 & 1.18 & 1.19 & 1.22 & 1.08 & 1.08 \\
SVHN & 1.12 & 1.60 & 1.61 & 1.65 & 1.51 & 1.51 \\
EuroSAT & 1.31 & 3.22 & 3.22 & 3.26 & 2.84 & 2.84 \\
\bottomrule
\end{tabular}
\end{center}
\end{table*}

\section{Larger Backbones} \label{app:scale}

\begin{table}[h]
\caption{ImageNet-100 at $224{\times}224$ with a fine-tuned ConvNeXt-T. Top-1 accuracy (\%) on the 5,000 validation images at the final epoch, mean$\pm$std.}\label{tab:in100}
\begin{center}\footnotesize
\setlength{\tabcolsep}{4.8pt}
\begin{tabular}{lcccccc}
\toprule
 & Unconstrained & $K{=}25$ & $K{=}50$ & $K{=}100$ & $K{=}150$ & $K{=}200$ \\
\midrule
ConvNeXt-T & 94.6$\pm$0.1 & 91.7$\pm$0.5 & 92.0$\pm$0.3 & 91.9$\pm$0.7 & 92.5$\pm$0.2 & 92.0$\pm$0.7 \\
\bottomrule
\end{tabular}
\end{center}
\end{table}

Table~\ref{tab:in100} reports ImageNet-100 with ConvNeXt-T, and Table~\ref{tab:in1k} reports ImageNet-1k with ConvNeXt-T. Experimental details for these runs can be found in Appendix~\ref{app:ExperimentalDetails}.

\begin{table}[h]
\caption{ImageNet-1k with a fine-tuned ConvNeXt-T. Top-1 accuracy (\%) on the 50,000 validation images at the final epoch, mean$\pm$std over two seeds.}\label{tab:in1k}
\begin{center}\footnotesize
\begin{tabular}{lcc}
\toprule
 & Top-1 & Seeds \\
\midrule
Unconstrained, 256-d linear projection & 81.2$\pm$0.0 & 2 \\
Unconstrained, 128-d linear projection & 81.1$\pm$0.1 & 2 \\
PNN, $K=128$ & 78.6$\pm$0.2 & 2 \\
PNN, $K=256$ & 78.9$\pm$0.1 & 2 \\
\bottomrule
\end{tabular}
\end{center}
\end{table}

\section{Additional Experimental Results}\label{sec:additional_experiments}

This appendix provides comprehensive experimental results and implementation details to support reproducibility. We organize the material as follows: Section~\ref{app:mlp} presents additional MLP classifier results; Section~\ref{app:cnn} provides CNN classifier analysis and visualizations; Section~\ref{app:ae} reports autoencoder reconstruction experiments.

\subsection{Multi-Layer Perceptron Classifiers}
\label{app:mlp}

\subsubsection{Tabular Dataset Performance}

We evaluate PNN-constrained MLPs on four tabular benchmark datasets: Iris, Wine, Breast Cancer, and Digits. These datasets vary in dimensionality (4-64 features), sample size (150-5,620 instances), and task complexity (2-10 classes), enabling systematic assessment of polytopal constraints under diverse conditions.

Table~\ref{tab:pca_vq_pnn_full} presents the mean test accuracy as a function of the number of archetypes $K$ for each dataset. Across all benchmarks, PNN-MLP approaches the performance of the unconstrained baseline as $K$ increases, with minimal degradation even at low archetype counts. Notably, PNN-MLP consistently outperforms both PNN-VQ-MLP and PNN-PL-MLP variants, demonstrating that soft convex combinations better preserve discriminative information compared to hard assignments (VQ) or unconstrained projections (PL). The Wine and Breast Cancer datasets exhibit particularly strong performance, achieving near-baseline accuracy with as few as 5 archetypes, suggesting that the decision boundaries in these tasks align well with polytopal geometry.

\subsubsection{Validation of Scalable Training Procedures}

To validate our scalable training framework (\ref{subsection:corpus_updates}), we compare mini-batched updates against full-batch optimization on three small datasets where exact computation is tractable. Table~\ref{tab:b_nb_accuracy} reports test accuracy for three projection solvers: CVX (generic convex solver), QP (quadratic programming with linear constraints), and SMO (our Sequential Minimal Optimization approach).

\begin{table*}[ht]
\caption{Test accuracy (\%) comparison between batched and non-batched training across optimization methods. Results show mean $\pm$ std. Batched training with SMO achieves statistically equivalent performance to exact methods while enabling scalability.}
\centering
\begin{footnotesize}
\begin{sc}
\begin{tabular}{p{0.1cm} c cc | cc | cc | cc}
\toprule
 & $K$
 & \multicolumn{2}{c}{CVX}
 & \multicolumn{2}{c}{QP}
 & \multicolumn{2}{c}{SMO}
 & \multicolumn{2}{c}{Amortized} \\
\cmidrule(lr){3-4} \cmidrule(lr){5-6} \cmidrule(lr){7-8} \cmidrule(lr){9-10}
 & 
 & Batch & $\neg$ Batch
 & Batch & $\neg$ Batch
 & Batch & $\neg$ Batch
 & Batch & $\neg$ Batch \\
\midrule
 \multirow{3}{*}{\rotatebox{90}{\textbf{Breast}}}
 & 5
 & 96.1$\pm$2.3 & 97.1$\pm$1.3
 & 96.4$\pm$1.5 & 96.4$\pm$1.8
 & 95.4$\pm$4.1 & 96.8$\pm$1.5
 & 96.3$\pm$2.2 & 96.3$\pm$1.8 \\
  
 & 10
 & 97.3$\pm$1.2 & 97.1$\pm$1.2
 & 96.4$\pm$1.2 & 97.2$\pm$1.4
 & 96.7$\pm$1.1 & 97.0$\pm$1.2
 & 96.7$\pm$1.7 & 96.3$\pm$1.7 \\
  
 & 25
 & 97.4$\pm$1.0 & 96.9$\pm$1.5
 & 96.5$\pm$0.0 & 97.4$\pm$2.0
 & 96.9$\pm$2.5 & 96.9$\pm$1.5
 & 96.0$\pm$0.7 & 96.3$\pm$1.4 \\
\midrule
 \multirow{3}{*}{\rotatebox{90}{\textbf{Iris}}}
 & 5
 & 93.7$\pm$4.0 & 94.7$\pm$2.8
 & 94.3$\pm$3.5 & 95.0$\pm$3.6
 & 93.7$\pm$4.0 & 93.0$\pm$5.1
 & 94.0$\pm$5.3 & 84.7$\pm$15.1 \\
  
 & 10
 & 95.3$\pm$2.8 & 95.3$\pm$2.8
 & 95.3$\pm$2.8 & 95.3$\pm$2.8
 & 95.7$\pm$2.7 & 96.0$\pm$3.1
 & 94.0$\pm$3.9 & 94.0$\pm$3.9 \\
  
 & 25
 & 95.0$\pm$3.2 & 95.0$\pm$3.2
 & 95.3$\pm$3.2 & 95.3$\pm$2.8
 & 94.7$\pm$2.8 & 95.3$\pm$2.8
 & 93.3$\pm$4.7 & 94.0$\pm$3.9 \\
\midrule
 \multirow{3}{*}{\rotatebox{90}{\textbf{Wine}}}
 & 5
 & 98.3$\pm$2.3 & 97.5$\pm$3.3
 & 98.3$\pm$2.3 & 97.2$\pm$3.2
 & 98.1$\pm$2.3 & 97.2$\pm$3.2
 & 97.2$\pm$1.8 & 97.8$\pm$2.1 \\
  
 & 10
 & 98.1$\pm$2.3 & 98.3$\pm$2.3
 & 97.5$\pm$2.4 & 97.5$\pm$2.4
 & 97.2$\pm$2.3 & 98.1$\pm$2.3
 & 98.9$\pm$1.4 & 98.3$\pm$2.2 \\
  
 & 25
 & 100.0$\pm$0.0 & 98.6$\pm$2.0
 & 98.6$\pm$2.0 & 98.6$\pm$2.0
 & 97.2$\pm$0.0 & 98.6$\pm$2.0
 & 98.3$\pm$2.2 & 97.8$\pm$2.1 \\
\bottomrule
\end{tabular}
\end{sc}
\end{footnotesize}
\label{tab:b_nb_accuracy}
\end{table*}

The results demonstrate that: (1) mini-batched training achieves equivalent accuracy to full-batch methods, validating our corpus streaming approach; (2) SMO-based updates match the performance of exact QP solvers while avoiding cubic-complexity matrix inversions; and (3) the simplified gradient computation (treating $\mathbf{S}$ as constant) introduces negligible bias. These findings confirm that our scalability optimizations preserve model quality while enabling application to large-scale datasets.

\subsubsection{Training Phase Ablation}

Table~\ref{tab:pca_vq_pnn_full} decomposes the performance across the three training phases described in \ref{app:trainingprotocol}: Pretrained (Phase 1, backbone only), Polytope (Phase 2, polytope only), and Full (Phase 3, end-to-end). 

\begin{table}[t]
  \caption{MLP test classification accuracy (\%) across training phases for PNN variants}
  \label{tab:pca_vq_pnn_full}
  \begin{center}
    \begin{footnotesize}
      \begin{sc}
        \begin{tabular}{p{0.2cm} l c |c c|c c|c c}
          \toprule
          & K & Pretrained & \multicolumn{2}{c|}{PNN-PL-MLP} & \multicolumn{2}{c|}{PNN-VQ-MLP} & \multicolumn{2}{c}{PNN-MLP} \\
          & & & Polytope & Full & Polytope & Full & Polytope & Full \\
          \midrule
          \multirow{7}{*}{\rotatebox{90}{\textbf{Breast}}} & 3 & 96.3$\pm$2.3 & 96.3$\pm$2.3 & 96.0$\pm$2.9 & 86.0$\pm$13.9 & 85.6$\pm$13.7 & 85.5$\pm$15.6 & 92.0$\pm$10.5 \\
           & 5 & 96.3$\pm$2.3 & 96.5$\pm$2.2 & 96.1$\pm$2.6 & 93.3$\pm$4.7 & 93.0$\pm$4.6 & 97.1$\pm$2.0 & 96.8$\pm$1.9 \\
           & 7 & 96.3$\pm$2.3 & 96.3$\pm$2.3 & 96.3$\pm$2.3 & 89.5$\pm$6.4 & 89.3$\pm$6.4 & 96.2$\pm$2.0 & 95.7$\pm$2.6 \\
           & 10 & 96.3$\pm$2.3 & 96.3$\pm$2.3 & 96.3$\pm$2.3 & 93.2$\pm$3.4 & 92.6$\pm$3.1 & 96.9$\pm$2.2 & 96.8$\pm$1.9 \\
           & 15 & 96.3$\pm$2.3 & 96.3$\pm$2.3 & 96.3$\pm$2.3 & 94.4$\pm$1.8 & 92.6$\pm$5.1 & 97.0$\pm$1.9 & 96.9$\pm$2.1 \\
           & 20 & 96.3$\pm$2.3 & 96.3$\pm$2.3 & 96.1$\pm$2.6 & 94.2$\pm$3.0 & 93.9$\pm$2.5 & 97.4$\pm$1.8 & 96.2$\pm$1.7 \\
           & 25 & 96.3$\pm$2.3 & 96.3$\pm$2.3 & 96.1$\pm$2.6 & 95.8$\pm$1.7 & 94.0$\pm$2.7 & 96.7$\pm$1.7 & 96.3$\pm$1.4 \\
          \midrule
          \multirow{7}{*}{\rotatebox{90}{\textbf{Digits}}} & 3 & 93.5$\pm$1.5 & 74.4$\pm$5.3 & 85.6$\pm$3.0 & 18.2$\pm$8.3 & 19.6$\pm$9.5 & 19.2$\pm$7.8 & 23.8$\pm$11.5 \\
           & 5 & 93.5$\pm$1.5 & 91.9$\pm$1.9 & 92.9$\pm$2.0 & 15.7$\pm$6.0 & 18.7$\pm$7.8 & 35.7$\pm$14.1 & 45.0$\pm$15.9 \\
           & 7 & 93.5$\pm$1.5 & 93.3$\pm$1.6 & 93.6$\pm$1.8 & 22.1$\pm$10.2 & 23.7$\pm$8.7 & 48.9$\pm$14.9 & 59.3$\pm$12.2 \\
           & 10 & 93.5$\pm$1.5 & 93.5$\pm$1.5 & 93.7$\pm$2.2 & 20.6$\pm$7.6 & 25.8$\pm$4.0 & 54.0$\pm$17.9 & 64.0$\pm$18.6 \\
           & 15 & 93.5$\pm$1.5 & 93.5$\pm$1.5 & 93.8$\pm$2.1 & 30.2$\pm$8.4 & 30.7$\pm$9.0 & 77.1$\pm$4.0 & 80.7$\pm$3.5 \\
           & 20 & 93.5$\pm$1.5 & 93.5$\pm$1.5 & 93.7$\pm$2.2 & 26.9$\pm$13.7 & 31.9$\pm$10.4 & 74.4$\pm$14.9 & 81.4$\pm$8.0 \\
           & 25 & 93.5$\pm$1.5 & 93.4$\pm$1.5 & 93.8$\pm$2.1 & 40.7$\pm$16.3 & 44.1$\pm$13.4 & 83.3$\pm$6.1 & 86.7$\pm$4.9 \\
          \midrule
          \multirow{7}{*}{\rotatebox{90}{\textbf{Iris}}} & 3 & 94.0$\pm$4.3 & 94.7$\pm$3.8 & 94.7$\pm$3.8 & 85.3$\pm$11.4 & 85.3$\pm$11.4 & 83.0$\pm$12.1 & 89.0$\pm$9.2 \\
           & 5 & 94.0$\pm$4.3 & 94.7$\pm$3.8 & 94.7$\pm$3.8 & 85.3$\pm$7.7 & 85.3$\pm$7.7 & 90.0$\pm$13.0 & 93.3$\pm$5.2 \\
           & 7 & 94.0$\pm$4.3 & 94.0$\pm$4.3 & 94.7$\pm$3.8 & 92.7$\pm$4.3 & 92.7$\pm$4.3 & 90.3$\pm$9.2 & 93.7$\pm$5.1 \\
           & 10 & 94.0$\pm$4.3 & 94.0$\pm$4.3 & 94.7$\pm$3.8 & 88.7$\pm$1.8 & 88.7$\pm$1.8 & 94.3$\pm$4.2 & 94.3$\pm$4.2 \\
           & 15 & 94.0$\pm$4.3 & 94.0$\pm$4.3 & 94.7$\pm$3.8 & 92.0$\pm$4.5 & 92.0$\pm$4.5 & 93.3$\pm$5.4 & 93.0$\pm$6.0 \\
           & 20 & 94.0$\pm$4.3 & 94.0$\pm$4.3 & 94.0$\pm$4.3 & 93.3$\pm$4.1 & 93.3$\pm$4.1 & 94.0$\pm$4.7 & 93.0$\pm$4.0 \\
           & 25 & 94.0$\pm$4.3 & 94.0$\pm$4.3 & 94.0$\pm$4.3 & 92.7$\pm$4.3 & 93.3$\pm$4.7 & 93.3$\pm$5.2 & 94.0$\pm$5.4 \\
          \midrule
          \multirow{7}{*}{\rotatebox{90}{\textbf{Wine}}} & 3 & 97.8$\pm$2.3 & 97.2$\pm$2.0 & 97.8$\pm$2.3 & 71.1$\pm$19.1 & 71.1$\pm$19.1 & 85.8$\pm$16.1 & 88.6$\pm$14.7 \\
           & 5 & 97.8$\pm$2.3 & 97.8$\pm$2.3 & 97.8$\pm$2.3 & 91.7$\pm$5.2 & 91.1$\pm$4.6 & 97.2$\pm$2.6 & 97.5$\pm$1.6 \\
           & 7 & 97.8$\pm$2.3 & 97.8$\pm$2.3 & 97.8$\pm$2.3 & 97.8$\pm$3.6 & 97.8$\pm$3.6 & 97.8$\pm$3.2 & 97.5$\pm$2.4 \\
           & 10 & 97.8$\pm$2.3 & 97.8$\pm$2.3 & 97.8$\pm$2.3 & 96.1$\pm$2.5 & 95.0$\pm$2.3 & 98.1$\pm$1.9 & 98.3$\pm$1.4 \\
           & 15 & 97.8$\pm$2.3 & 97.8$\pm$2.3 & 98.3$\pm$2.5 & 96.7$\pm$3.0 & 97.2$\pm$2.8 & 97.2$\pm$2.6 & 96.9$\pm$3.1 \\
           & 20 & 97.8$\pm$2.3 & 97.8$\pm$2.3 & 98.3$\pm$2.5 & 96.7$\pm$3.6 & 97.2$\pm$3.9 & 96.9$\pm$2.8 & 97.2$\pm$2.6 \\
           & 25 & 97.8$\pm$2.3 & 97.8$\pm$2.3 & 98.3$\pm$2.5 & 97.8$\pm$3.0 & 97.8$\pm$3.0 & 98.1$\pm$2.3 & 98.1$\pm$2.3 \\
          \midrule
          \bottomrule
        \end{tabular}
      \end{sc}
    \end{footnotesize}
  \end{center}
  \vskip -0.1in
\end{table}

Several patterns emerge from this ablation. First, for standard PNN-MLP, the Polytope phase stays within 4 points of the Pretrained accuracy on Breast Cancer, Iris and Wine for $K\ge5$, but reaches only 35.7 to 83.3\% on Digits.

Second, PNN-PL consistently maintains pretrained performance in Phase 2, confirming that unconstrained subspace projections impose negligible information loss. The near-perfect preservation (e.g., 97.8\% on Wine in the Polytope phase for every $K\ge5$) demonstrates that the learned subspaces capture the essential structure of the pretrained features.

Third, PNN-VQ exhibits substantial performance degradation in Phase 2, particularly at low $K$ values. For the Breast Cancer dataset with $K=3$, PNN-VQ drops to 86.0\% compared to the 96.3\% pretrained baseline, indicating that hard discrete assignments cannot faithfully represent the continuous structure of pretrained features without co-adaptation. However, the performance recovers somewhat as $K$ increases, suggesting that a sufficiently large codebook can approximate the required representational capacity. PNN-VQ-MLP stays competitive on Breast Cancer, Iris and Wine at larger $K$.

Finally, Phase 3 end-to-end training (Full) provides modest but consistent improvements for standard PNN. This validates our three-phase protocol: pretraining establishes a strong initialization, polytope-only optimization verifies representational capacity, and joint fine-tuning enables full co-adaptation when necessary.

\subsection{CNN Experiments}\label{app:cnn}

\begin{table}[ht]
\caption{CNN test accuracy (\%) on three MedMNIST datasets at $64{\times}64$, mean$\pm$std over five seeds. Each PNN run starts from the unconstrained network of the same seed and uses the deployed coordinate map.}\label{tab:medmnist}
\begin{center}\footnotesize
\begin{tabular}{lcccc}
\toprule
 & Classes & Unconstrained & PNN $K=10$ & PNN $K=25$ \\
\midrule
PathMNIST & 9 & 90.4$\pm$1.2 & 88.5$\pm$1.5 & 89.1$\pm$1.5 \\
DermaMNIST & 7 & 78.6$\pm$0.7 & 76.4$\pm$0.6 & 77.4$\pm$0.7 \\
BloodMNIST & 8 & 98.3$\pm$0.0 & 98.1$\pm$0.2 & 98.1$\pm$0.1 \\
\bottomrule
\end{tabular}
\end{center}
\end{table}

Table~\ref{tab:medmnist} applies the PNN-CNN to PathMNIST, DermaMNIST and BloodMNIST from MedMNIST v2 \citep{Yang2023MedMNISTClassification}, with $K\in\{10,25\}$ and five seeds, more information in Appendix~\ref{app:ExperimentalDetails}.

\begin{table}[!ht]
\caption{CNN test accuracy (\%) with one constrained site or two, deployed coordinates (the amortizer followed by $K$ rounds of warm-started batched SMO), mean$\pm$std over the same three seeds for every column. The configuration is otherwise identical.}\label{tab:layercount}
\begin{center}\footnotesize
\begin{tabular}{llccc}
\toprule
 & $K$ & Unconstrained & One site & Two sites \\
\midrule
MNIST & 3 & 99.6$\pm$0.0 & 98.4$\pm$1.3 & 91.7$\pm$12.8 \\
 & 10 &  & 99.4$\pm$0.0 & 99.4$\pm$0.1 \\
 & 25 &  & 99.5$\pm$0.0 & 99.5$\pm$0.0 \\
FMNIST & 3 & 93.2$\pm$0.1 & 92.1$\pm$0.1 & 91.6$\pm$0.6 \\
 & 10 &  & 92.8$\pm$0.3 & 92.8$\pm$0.2 \\
 & 25 &  & 93.0$\pm$0.1 & 92.7$\pm$0.1 \\
CIFAR-10 & 3 & 82.3$\pm$0.3 & 76.9$\pm$0.7 & 75.7$\pm$0.6 \\
 & 10 &  & 81.6$\pm$0.4 & 81.5$\pm$0.5 \\
 & 25 &  & 81.8$\pm$0.4 & 82.2$\pm$0.4 \\
\bottomrule
\end{tabular}
\end{center}
\end{table}

Table~\ref{tab:layercount} compares one constrained site with two under an otherwise identical configuration on MNIST, FashionMNIST and CIFAR-10.

\subsection{AE Experiments}\label{app:ae}

\begin{figure}[htbp] 
    
    \centering
    \begin{subfigure}[b]{0.48\textwidth}
        \centering
        \includegraphics[width=\linewidth]{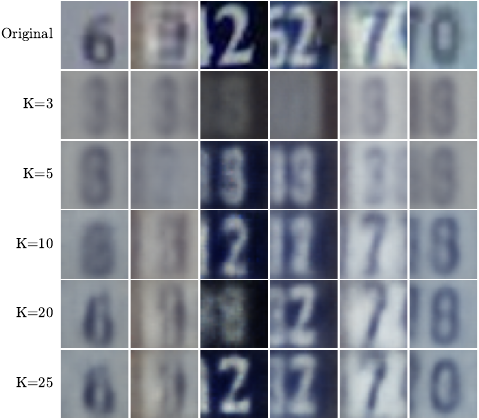}
        \caption{SVHN}
    \end{subfigure}
    \hfill 
    \begin{subfigure}[b]{0.48\textwidth}
        \centering
        \includegraphics[width=\linewidth]{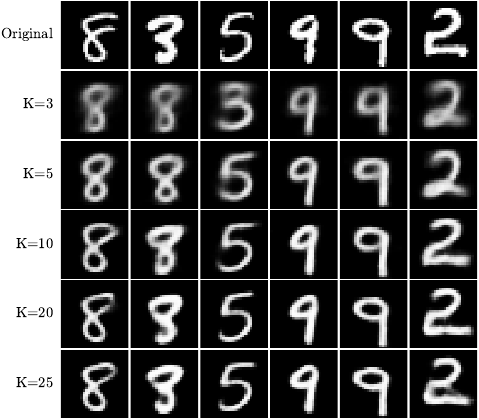}
        \caption{MNIST}
    \end{subfigure}
    
    \vspace{1em} 
    
    \begin{subfigure}[b]{0.48\textwidth}
        \centering
        \includegraphics[width=\linewidth]{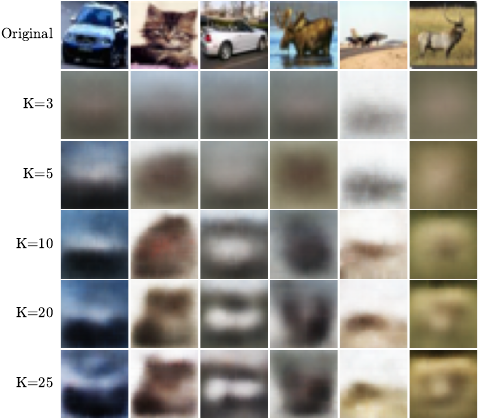}
        \caption{CIFAR-10}
    \end{subfigure}
    \hfill 
    \begin{subfigure}[b]{0.48\textwidth}
        \centering
        \includegraphics[width=\linewidth]{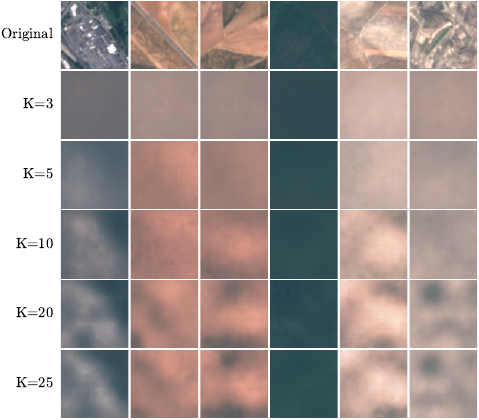}
        \caption{EuroSAT}
    \end{subfigure}
    
    \caption{PNN-AE reconstruction by K for SVHN, MNIST, CIFAR-10, EuroSAT}
    \label{fig:recon_by_k}
\end{figure}

From Figure~\ref{fig:recon_by_k} we see how the size of the polytope affects the quality of the reconstruction in the datasets. A larger polytope means less compression and therefore more expressive power which results in better reconstruction.

\begin{figure}[htbp] 
    \centering
    \begin{subfigure}[b]{0.48\textwidth}
        \centering
        \includegraphics[width=\linewidth]{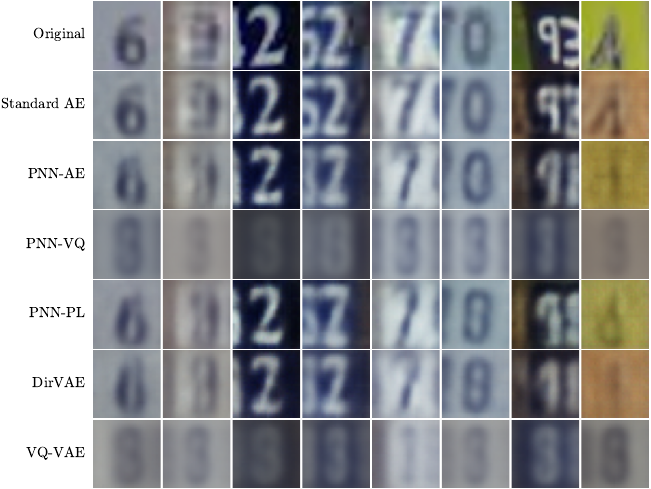}
        \caption{SVHN}

    \end{subfigure}
    \hfill 
    \begin{subfigure}[b]{0.48\textwidth}
        \centering
        \includegraphics[width=\linewidth]{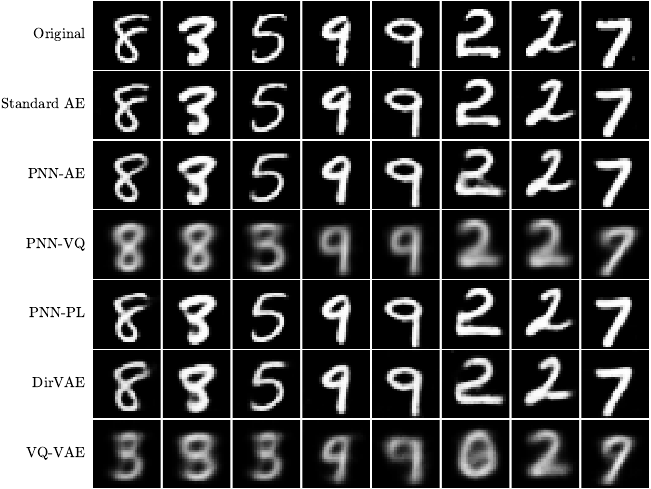}
        \caption{MNIST}
    \end{subfigure}
    
    \vspace{1em} 
    
    \begin{subfigure}[b]{0.48\textwidth}
        \centering
        \includegraphics[width=\linewidth]{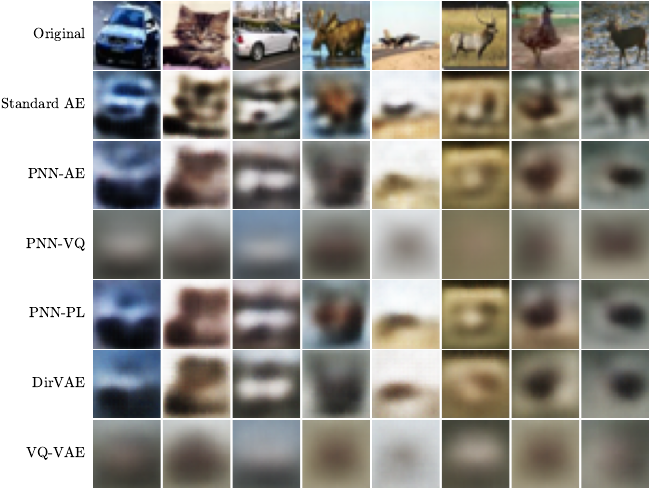}
        \caption{CIFAR-10}
    \end{subfigure}
    \hfill 
    \begin{subfigure}[b]{0.48\textwidth}
        \centering
        \includegraphics[width=\linewidth]{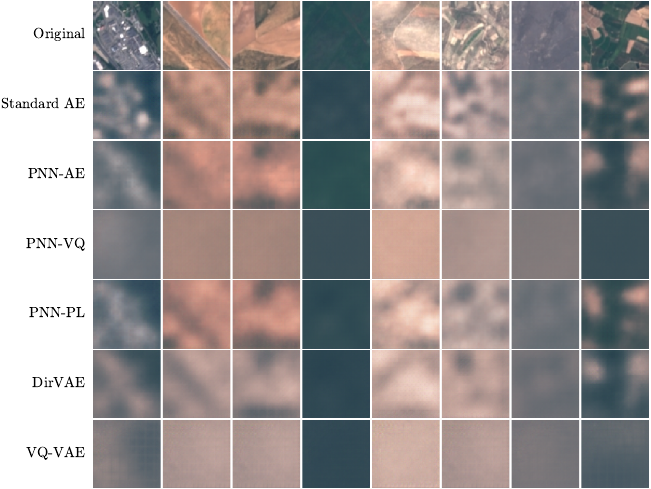}
        \caption{EuroSAT}
    \end{subfigure}
    
    \caption{Reconstruction by model for $K=25$ for SVHN, MNIST, CIFAR-10 and EuroSAT.}
    \label{fig:recon_by_model}
\end{figure}

From Figure~\ref{fig:recon_by_model} shows how each of the models reconstruct across datasets. We see that the PNN visually offers competitive reconstruction quality. 

In table \ref{tab:sparsemax} we further compared using a softmax instead of a sparsemax in the amortizer. Both methods have similar mean MSE. 
\begin{table}[ht]
\caption{PNN-AE reconstruction MSE with a softmax or sparsemax amortizer output. Values are mean$\pm$std over the same three seeds.}\label{tab:sparsemax}
\begin{center}\footnotesize
\begin{tabular}{llcc}
\toprule
 & $K$ & Softmax & Sparsemax \\
\midrule
MNIST & 3 & 0.0399$\pm$0.0003 & 0.0400$\pm$0.0005 \\
 & 10 & 0.0147$\pm$0.0005 & 0.0148$\pm$0.0001 \\
 & 25 & 0.0073$\pm$0.0001 & 0.0075$\pm$0.0003 \\
FMNIST & 3 & 0.0266$\pm$0.0002 & 0.0268$\pm$0.0003 \\
 & 10 & 0.0127$\pm$0.0001 & 0.0127$\pm$0.0001 \\
 & 25 & 0.0099$\pm$0.0002 & 0.0100$\pm$0.0004 \\
CIFAR-10 & 3 & 0.0363$\pm$0.0001 & 0.0363$\pm$0.0001 \\
 & 10 & 0.0218$\pm$0.0000 & 0.0218$\pm$0.0000 \\
 & 25 & 0.0152$\pm$0.0004 & 0.0157$\pm$0.0007 \\
\bottomrule
\end{tabular}
\end{center}
\end{table}

\subsubsection{Vector Quantization}\label{app:AppendixVQ}

To benchmark polytopal soft-clustering against discrete hard-clustering, we compare to the VQ-VAE. VQ-VAE discretizes the latent space using a finite codebook of embeddings $\mathbf{E}^\ell$.

The codebook is updated using an exponential moving average (EMA) scheme rather than direct gradient descent. Gradients are propagated through the quantization operation using the Straight-Through Estimator (STE), passing gradients from quantized latents $\mathbf{R}^\ell$ to encoder outputs $\mathbf{Z}^\ell$ unchanged.

The training objective is
\begin{equation}
\mathcal{L} = \mathcal{L}_{\text{task}} + \beta \sum_{\ell} ||\mathbf{Z}^\ell - \operatorname{sg}[\mathbf{R}^\ell]||_F^2,
\end{equation}
where the commitment term encourages encoder outputs to remain close to their assigned codebook vectors. To prevent codebook collapse, $\mathbf{E}^\ell$ is initialized via greedy k-means++ on the full training set, and unused embeddings are periodically reinitialized.

A direct comparison between the PNN-VQ and the VQ-VAE can be seen in table~\ref{tab:vq-full}.
\begin{table}[ht]
\caption{Reconstruction MSE of a standard VQ-VAE (EMA codebook, commitment loss, straight-through estimator) and of PNN-VQ (direct gradient updates of a corpus-grounded codebook, reconstruction loss only), mean$\pm$std over five seeds.}\label{tab:vq-full}
\begin{center}\footnotesize
\begin{tabular}{llcc}
\toprule
 & $K$ & VQ-VAE & PNN-VQ \\
\midrule
MNIST & 3 & 0.0613$\pm$0.0005 & 0.0614$\pm$0.0006 \\
 & 5 & 0.0568$\pm$0.0006 & 0.0580$\pm$0.0007 \\
 & 10 & 0.0530$\pm$0.0006 & 0.0527$\pm$0.0009 \\
 & 15 & 0.0502$\pm$0.0005 & 0.0493$\pm$0.0004 \\
 & 20 & 0.0478$\pm$0.0001 & 0.0474$\pm$0.0003 \\
 & 25 & 0.0464$\pm$0.0004 & 0.0458$\pm$0.0007 \\
\midrule
FMNIST & 3 & 0.0635$\pm$0.0049 & 0.0611$\pm$0.0013 \\
 & 5 & 0.0540$\pm$0.0008 & 0.0542$\pm$0.0010 \\
 & 10 & 0.0459$\pm$0.0014 & 0.0431$\pm$0.0005 \\
 & 15 & 0.0408$\pm$0.0008 & 0.0394$\pm$0.0005 \\
 & 20 & 0.0381$\pm$0.0008 & 0.0370$\pm$0.0004 \\
 & 25 & 0.0362$\pm$0.0004 & 0.0361$\pm$0.0003 \\
\midrule
CIFAR-10 & 3 & 0.0552$\pm$0.0041 & 0.0477$\pm$0.0004 \\
 & 5 & 0.0452$\pm$0.0004 & 0.0444$\pm$0.0003 \\
 & 10 & 0.0412$\pm$0.0005 & 0.0414$\pm$0.0005 \\
 & 15 & 0.0393$\pm$0.0001 & 0.0395$\pm$0.0004 \\
 & 20 & 0.0382$\pm$0.0002 & 0.0383$\pm$0.0001 \\
 & 25 & 0.0374$\pm$0.0002 & 0.0376$\pm$0.0002 \\
\midrule
SVHN & 3 & 0.0339$\pm$0.0025 & 0.0262$\pm$0.0004 \\
 & 5 & 0.0278$\pm$0.0032 & 0.0224$\pm$0.0004 \\
 & 10 & 0.0279$\pm$0.0058 & 0.0207$\pm$0.0008 \\
 & 15 & 0.0221$\pm$0.0037 & 0.0197$\pm$0.0002 \\
 & 20 & 0.0186$\pm$0.0011 & 0.0190$\pm$0.0003 \\
 & 25 & 0.0172$\pm$0.0002 & 0.0181$\pm$0.0003 \\
\midrule
EuroSAT & 3 & 0.0145$\pm$0.0007 & 0.0120$\pm$0.0005 \\
 & 5 & 0.0106$\pm$0.0007 & 0.0107$\pm$0.0006 \\
 & 10 & 0.0095$\pm$0.0003 & 0.0096$\pm$0.0005 \\
 & 15 & 0.0089$\pm$0.0001 & 0.0092$\pm$0.0002 \\
 & 20 & 0.0088$\pm$0.0002 & 0.0090$\pm$0.0001 \\
 & 25 & 0.0085$\pm$0.0000 & 0.0091$\pm$0.0003 \\
\bottomrule
\end{tabular}
\end{center}
\end{table}

\end{document}